\documentclass[acmtog, authorversion, nonacm]{acmart} % for arxiv
\acmSubmissionID{437}

\usepackage{booktabs} % For formal tables
\usepackage{tabularx} % for 'tabularx' environment and 'X' column type
\usepackage{xcolor}
\usepackage{euscript} % for curly fonts
\usepackage{bbm} % for indicator function 1
\usepackage[ruled]{algorithm2e} % For algorithms

\SetAlFnt{\small}
\SetAlCapFnt{\small}
\SetAlCapNameFnt{\small}
\SetAlCapHSkip{0pt}

\copyrightyear{2024}
\acmYear{2024}
\setcopyright{rightsretained}
\acmConference[SA Conference Papers '24]{SIGGRAPH Asia 2024 Conference
Papers}{December 3--6, 2024}{Tokyo, Japan}
\acmBooktitle{SIGGRAPH Asia 2024 Conference Papers (SA Conference Papers '24), December 3--6, 2024, Tokyo, Japan}
\acmDOI{10.1145/3680528.3687605}
\acmISBN{979-8-4007-1131-2/24/12}

\newif\ifdraft
\draftfalse

\newif\ifoverpic
\overpictrue
\newif\ifwithsupp
\withsupptrue
\ifdraft
\newcommand{\itai}[1]{{\color{magenta}[\textbf{Itai:} #1]}}
\newcommand{\dale}[1]{{\color{orange}[\textbf{Dale:} #1]}}
\newcommand{\sudarhan}[1]{{\color{purple}[\textbf{Sudarshan:} #1]}}
\newcommand{\rana}[1]{{\color{cyan}[\textbf{Rana:} #1]}}

\newcommand{\camrdy}[1]{{\color{blue}#1}}
\newcommand{\siga}[1]{{\color{black}#1}}
\newcommand{\old}[1]{{\color{red}\sout{#1}}}

\else
\newcommand{\itai}[1]{}
\newcommand{\dale}[1]{}
\newcommand{\sudarhan}[1]{}
\newcommand{\rana}[1]{}
\newcommand{\camrdy}[1]{{\color{black}#1}}
\newcommand{\siga}[1]{{\color{black}#1}}
\newcommand{\old}[1]{{\color{black}}}
\fi

\newcommand{\ourmethod}{iSeg}

\DeclareMathOperator*{\argmax}{\arg\!\max}

\newcommand{\Rcal}{\mathcal{R}}
\newcommand{\Lcal}{\mathcal{L}}
\newcommand{\Mcal}{\mathcal{M}}

\makeatletter
\DeclareRobustCommand\onedot{\futurelet\@let@token\@onedot}
\def\@onedot{\ifx\@let@token.\else.\null\fi\xspace}

\def\eg{\emph{e.g}\onedot}

\def\wrt{w.r.t\onedot}

\makeatother

\let\titleold\title
\renewcommand{\title}[1]{\titleold{#1}\newcommand{\thetitle}{#1}}
\def\maketitlesupplementary
   {
   \newpage
       \twocolumn[
        \centering
        \Large
        \textbf{\thetitle}\\
        \vspace{0.5em}Supplementary Material \\
        \vspace{1.0em}
       ] %< twocolumn
   }

\AtEndPreamble{
    \usepackage[capitalize]{cleveref}
    \crefname{section}{Sec.}{Secs.}
    \Crefname{section}{Section}{Sections}
    \Crefname{table}{Table}{Tables}
    \crefname{table}{Tab.}{Tabs.}
}

\RequirePackage{enumitem}
\setlist[itemize]{noitemsep,leftmargin=*,topsep=0em}
\setlist[enumerate]{noitemsep,leftmargin=*,topsep=0em}

\ifwithsupp
\usepackage{overpic} % for overlaying text over figures
\fi

\begin{document}
% Title portion
\title{\ourmethod{}: Interactive 3D Segmentation via Interactive Attention}

%%%%%%%%% AUTHORS
\author{Itai Lang}
\affiliation{
  \institution{University of Chicago}
  \country{United States of America}
}
\email{itailang@uchicago.edu}

\author{Fei Xu}
\affiliation{
  \institution{University of Chicago}
  \country{United States of America}
}
\email{feixu@uchicago.edu}

\author{Dale Decatur}
\affiliation{
  \institution{University of Chicago}
  \country{United States of America}
}
\email{ddecatur@uchicago.edu}

\author{Sudarshan Babu}
\affiliation{
  \institution{Toyota Technological Institute at Chicago}
  \country{United States of America}
}
\email{sudarshan@ttic.edu}

\author{Rana Hanocka}
\affiliation{
  \institution{University of Chicago}
  \country{United States of America}
}
\email{ranahanocka@uchicago.edu}

\renewcommand\shortauthors{Itai Lang, Fei Xu, Dale Decatur, Sudarshan Babu, and Rana Hanocka}

%%%%%%%%% ABSTRACT
%%%%%%%%% ABSTRACT
\begin{abstract}

We present \ourmethod{}, a new interactive technique for segmenting 3D shapes. Previous works have focused mainly on leveraging pre-trained 2D foundation models for 3D segmentation based on text. However, text may be insufficient for accurately describing fine-grained spatial segmentations. Moreover, achieving a consistent 3D segmentation using a 2D model is highly challenging, since occluded areas of the same semantic region may not be visible together from any 2D view. Thus, we design a segmentation method conditioned on fine user clicks, which operates entirely in 3D. Our system accepts user clicks directly on the shape's surface, indicating the inclusion or exclusion of regions from the desired shape partition. To accommodate various click settings, we propose a novel interactive attention module capable of processing different numbers and types of clicks, enabling the training of a single unified interactive segmentation model. We apply \ourmethod{} to a myriad of shapes from different domains, demonstrating its versatility and faithfulness to the user's specifications. Our project page is at \url{https://threedle.github.io/iSeg/}.

\end{abstract}




% Teaser
%\ifoverpic
%\input{figures/teaser/teaser_png.tex}
%\else
\begin{teaserfigure}
\begin{center}
    \centering
    % trim: left, bottom, right, top
    \includegraphics[width=\textwidth, trim=10 70 20 170, clip]{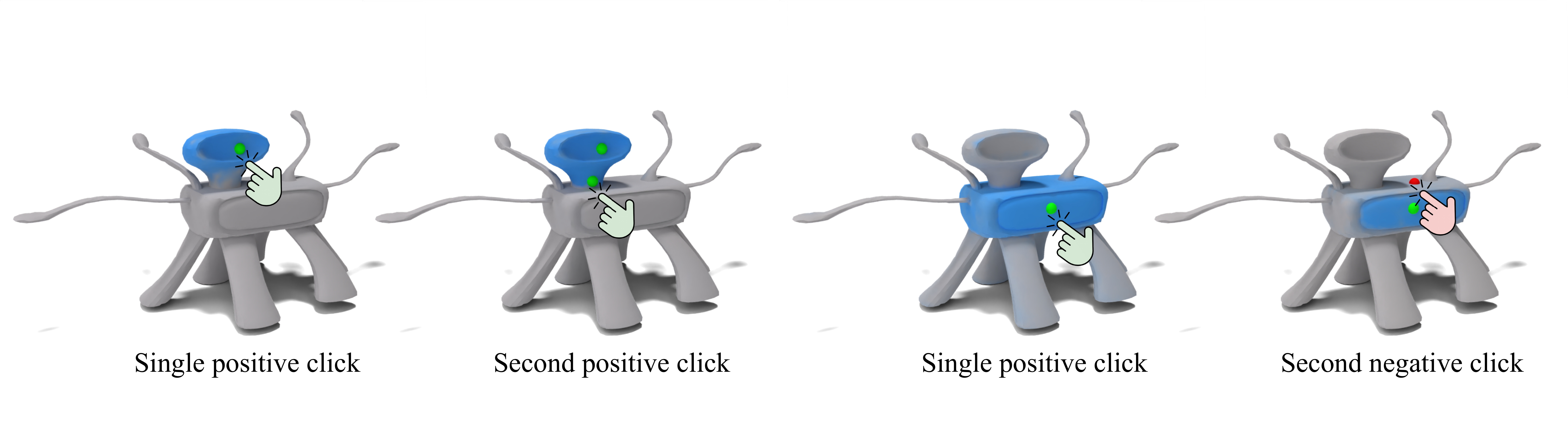}
    \vspace{-5mm}
    \caption{\ourmethod{} computes customized fine-grained segmentations on shapes interactively specified by user clicks. The clicks, denoted by a green or a red dot, indicate whether to include or exclude regions, respectively. Our method is capable of segmenting regions that are not accurately specified by text.}
    \vspace{2mm}
    \label{fig:teaser}
\end{center}
\end{teaserfigure}

%\fi

\maketitle

%%%%%%%%% BODY
\section{Introduction} \label{sec:introduction}

%%%%%%%% introduction to interactive segmentation problem & what we want to achieve & why it's hard (and how it's different from traditional segmentation)
% to mention: 3d modeling workflows, interactively select different regions, problem ill-defined (how do we even acquire the training data for it?) %%%%%%%%

%%%%%%%%%%%%%%%%%%%%%%%%%%%%%%%%%%%%%%%%%%%%%%%%%%%%
% Introducing the interactive segmentation problem %
%%%%%%%%%%%%%%%%%%%%%%%%%%%%%%%%%%%%%%%%%%%%%%%%%%%%
\siga{
Interactive 3D segmentation, the ability to select fine-grained segments from a 3D shape based on user inputs like clicks, is a fundamental problem in computer graphics with broad implications. In fields such as computer-aided design and 3D modeling, precise segment selection facilitates detailed model refinement. Moreover, in engineering, architecture, and medicine, fine-grained selection is indispensable for simulation and analysis, allowing for accurate assessment of structural integrity and behavior. While important, this problem poses significant challenges. How can we decipher the user intentions from such a minimal input as clicks? How do we handle diverse shapes with varying geometries and select specific and unique shape parts? In this work, we propose a method tailored to the shape at hand that selects regions adhering to the user clicks.
}

%%%%%%%%%%%%%%%%%%%%%%%%%%%%%%%%%%%%%%%%%
% evolution to data-driven segmentation %
%%%%%%%%%%%%%%%%%%%%%%%%%%%%%%%%%%%%%%%%%
Traditional segmentation techniques do not utilize user inputs and instead rely on geometric features to delineate semantic parts \cite{hoffman1984parts, dey2004approximating, lien2007approximate, cornea2007curve, shamir2008survey, skeleton_intrinsic}. Recent data-driven techniques have further leveraged fully annotated 3D datasets and achieved ipressive 3D segmentation results \cite{milletari2016vnet, qi2017pointnet, yi2017syncspeccnn, hanocka2019meshcnn, chen2019bae, subdivnet, zhu2020adacoseg, milano2020primal, deng2020cvxnet, sharp2022diffusionnet, canonical_capsules}. 
However, the reliance on a dataset and the scarcity of large-scale 3D datasets limits the network to a specific shape domain with a predefined set of parts.

%%%%%%%%%%%%%%%%%%%%%%%%%%%%%%%%%%%%%%%%%%%%%%%%%%
% evolution to zero-shot segmentation and the gap %
%%%%%%%%%%%%%%%%%%%%%%%%%%%%%%%%%%%%%%%%%%%%%%%%%%
Current 3D segmentation methods have circumvented the dependency on 3D data and pre-determined part definition by utilizing pretrained 2D foundation models to learn semantic co-segmentation~\cite{ye2023featurenerf} or text-driven segmentation~\cite{ha2022semabs, decatur2023highlighter, abdelreheem2023SATR, abdelreheem2023zero, liu2023partslip, decatur2024paintbrush, kim2024partstad}.
Nonetheless, text may not be able to accurately describe all fine-grained segmentations, such as the fourth leg of an octopus or a region corresponding to a particular point on the shape.

\begin{figure*}
\centering
\newcommand{\pl}{-2}
% trim: left, bottom, right, top
\includegraphics[width=\linewidth, trim=80 50 20 0, clip]{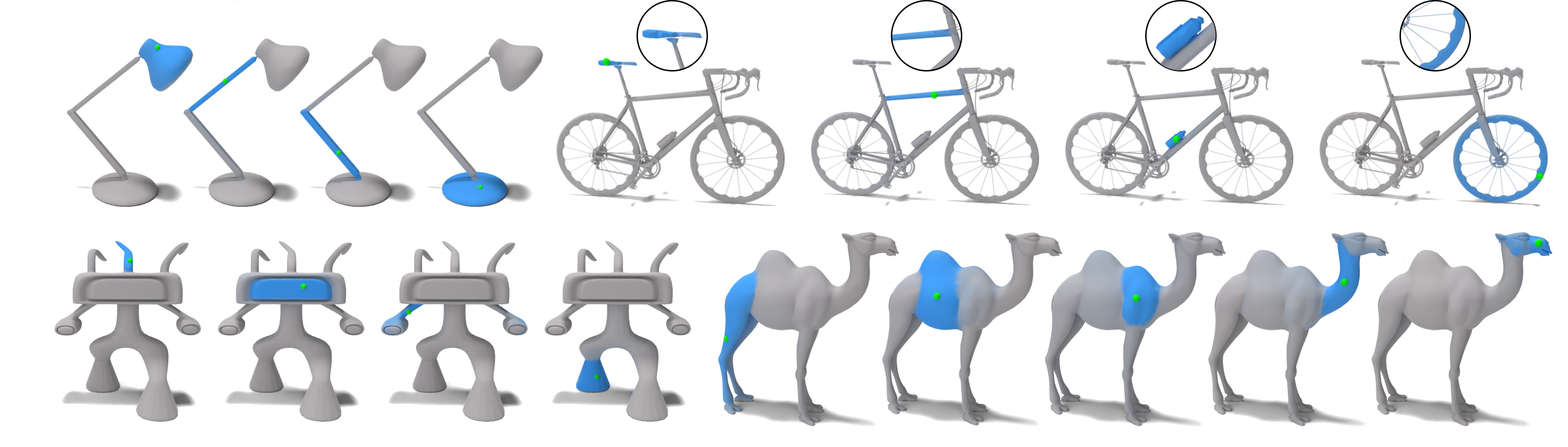}
\caption{\textbf{Fine-grained segmentation from a single positive click.} \ourmethod{} is capable of generating granular segmentations (visualized in blue) given a single click as input (depicted with a green dot). Our method is highly flexible and can select parts that vary in size, geometry, and semantic meaning.}
\vspace{-3pt}
\label{fig:single}
\end{figure*}

%%%%%%%%%%%%%%%%%%%%%%%%%%%%%%%%%%%%%%%
% What we did (our answer to the gap) %
%%%%%%%%%%%%%%%%%%%%%%%%%%%%%%%%%%%%%%%
In this paper, we present \textit{\ourmethod{}}, a new data-driven interactive technique for 3D shape segmentation that generates customized partitions of the shape according to user clicks.
Given a shape represented as a triangular mesh, the user selects points on the mesh interactively to indicate a desired segmentation and \ourmethod{} predicts a region over the mesh surface that adheres to the clicked points.
Our interactive interface can utilize positive and negative clicks, enabling additions and exclusions of areas from the segmented region, respectively (see \cref{fig:teaser}).

%%%%%%%%%%%%%%%%%%%%%%%%%%%%
% how we did it (overview) %
%%%%%%%%%%%%%%%%%%%%%%%%%%%%
% the goals of the paragraph: 
% (a) state the challenge of devising a meaningful 3D segmentation model from 2D supervision
% (b) explain how we overcame this challenge by designing a native 3D system and its benefits
We harness the power of a pretrained 2D foundation segmentation model \cite{kirillov2023segment} and distill its knowledge to 3D.
However, segmenting a meaningful 3D region using a 2D model is very challenging, since occluded shape regions cannot be seen together from a single 2D view.
Accordingly, we design an interactive segmentation system that operates \textit{entirely in 3D}, where the user clicks and the inferred corresponding region are applied over the shape surface \textit{directly}, ensuring 3D consistency \textit{by construction}.
During training only, we project the 3D user clicks and the predicted segmentation to multiple 2D views to enable supervision from the powerful pretrained foundation model \cite{kirillov2023segment}.

%%%%%%%%%%%%%%%%%%%%%%%%%%%
% how we did it (novelty) %
%%%%%%%%%%%%%%%%%%%%%%%%%%%
% the goals of this paragraph:
% (a) present our main contribution - interactive attention
% (b) explain its benefits - enable a unified model for various interactive user inputs
For interactiveness, we want our system to accommodate different user inputs, meaning, point clicks that can vary in number and type.
Instead of training a separate segmentation model for each user click configuration, we propose a novel interactive attention mechanism, which learns the representation of positive and negative clicks and computes their interaction with the other points of the mesh.
This attention layer consolidates variable-size guidance into a fixed-size representation, resulting in a unified flexible segmentation model capable of predicting shape regions for various click settings.

\camrdy{\ourmethod{} is optimized per mesh to capture its unique segments, without any ground-truth annotations. We train the model with only a small fraction of the mesh vertices, while the model successfully infers segmentations for other vertices not used during training. \ourmethod{} further generalizes beyond its training data and computes complete segments in 3D for clicks and regions occluded from each other.}

%%%%%%%%%%%%%%%%%%%%%%%%%
% summary + what we got %
%%%%%%%%%%%%%%%%%%%%%%%%%
In summary, this paper presents \ourmethod{}, an interactive method for selecting customized fine-grained regions on a 3D shape. We distill inconsistent feature embeddings of a 2D foundation model into a coherent feature field over the mesh surface and decode it along with user inputs to segment the mesh on the fly. Our interactive attention mechanism handles a variable number of user clicks that can signify both the inclusion and exclusion of regions. We showcase the effectiveness of \ourmethod{} on a variety of meshes from different domains, including humanoids, animals, and man-made objects, and show its flexibility for various segmentation specifications.

\section{Related Work} \label{sec:related_work}

\begin{figure*}
\centering
% trim: left, bottom, right, top
\includegraphics[width=0.99\linewidth, trim=0 10 30 0, clip]{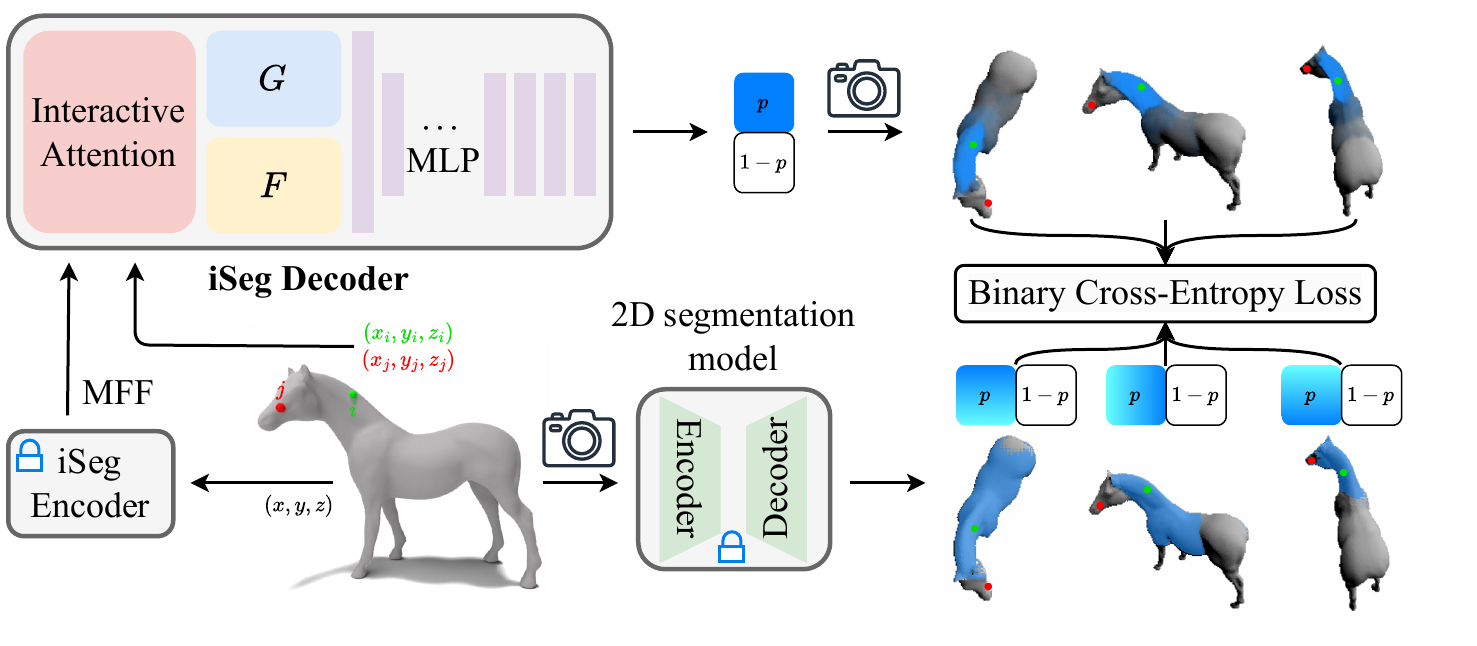}
\vspace{-0.3cm}
\caption{\textbf{Training of \siga{the} \ourmethod{} decoder.} Our decoder takes the Mesh Feature Field (MFF) \siga{computed by the} \ourmethod{} encoder, along with the user input clicks, and generates a 3D segmentation map visualized in blue. We leverage a pre-trained 2D segmentation model \cite{kirillov2023segment} to supervise our training with 2D segmentation masks using rendered images of the shape and the 2D projection of the 3D clicks. Although \ourmethod{} is trained using noisy and inconsistent 2D segmentations, it is view-consistent by construction.}
\label{fig:system}
\end{figure*}

%%%%%%%%%%%%%%%%%%%%%%%%%%%%%%%%%%%
% Non-interactive 3D Segmentation %
%%%%%%%%%%%%%%%%%%%%%%%%%%%%%%%%%%%
%\noindent \textbf{Non-interactive 3D segmentation.}
\subsection{Non-Interactive 3D Segmentation}
A large body of research has been focused on 3D segmentation using annotated datasets \cite{2017arXiv170201105A, hanocka2019meshcnn, subdivnet, milano2020primal, sharp2022diffusionnet, yi2017syncspeccnn, kalogerakis2017pcn}. Such models demonstrate impressive performance at the cost of being restricted to the domain of the training data and the set of manually defined semantic labels. A partial solution to this limitation is utilizing unlabeled data, where common semantic elements are discovered by unsupervised learning \cite{chen2019bae, deng2020cvxnet, hong2022threedcg, canonical_capsules, zhu2020adacoseg}. Still, the segmentation is confined to the learned parts and is not easy to alter.

In contrast, our segmentation approach is highly versatile and flexible. It is applied to various shapes from different domains. Our model is trained without any segment labels, and instead, it is optimized to the shape at hand to discover its unique partitions. Moreover, \ourmethod{} is interactive -- its segmentation result can be updated simply with an intuitive user-click interface.

%%%%%%%%%%%%%%%%%%%%%%%%%%%%
% Lifting 2D models to 3D  %
%%%%%%%%%%%%%%%%%%%%%%%%%%%%
%\smallskip
%\noindent \textbf{Lifting 2D foundation models to 3D.}
\subsection{Lifting 2D Foundation Models to 3D}
The emergence of powerful 2D foundation models with a broad semantic understanding has propelled a surge of interest in distilling their knowledge and lifting it to a 3D representation \cite{decatur2023highlighter, 2020arXiv200713138K, zhang2022learning, chen2023bridging, abdelreheem2023SATR, abdelreheem2023zero, peng2023openscene, fan2022nerf-sos, umam2024partdistill, yin2024sai3d, decatur2024paintbrush}. Notably, several researchers \cite{tschernezki22neural, kobayashi2022decomposing, kerr2023lerf} augmented the neural radiance scene representation (NeRF) \cite{mildenhall2020nerf} with a volumetric feature filed. This approach enabled text-driven segmentation of objects within the scene, alleviating the need for a training dataset.

Similarly, we lift the features of a 2D foundation model \cite{kirillov2023segment} into 3D. \siga{However, instead of using the implicit NeRF representation \cite{ye2023featurenerf, kerr2023lerf}, our model operates directly on explicit 3D meshes, making it readily adaptable to 3D modeling workflows. Moreover, rather than decoding the feature field by a simple correlation with the embedding of the semantic prompt \siga{\cite{fan2022nerf-sos, peng2023openscene}}, we learn a dedicated decoder in 3D to exploit the semantic information embodied within our mesh feature field better.}

%%%%%%%%%%%%%%%%%%%%%%%%%%%%%%%
% Interactive 3D Segmentation %
%%%%%%%%%%%%%%%%%%%%%%%%%%%%%%%
% We first talk about traditional techniques
%\smallskip
%\noindent \textbf{Interactive 3D segmentation.}
\subsection{Interactive 3D Segmentation}
Traditional interactive techniques have utilized heuristic smoothness priors and formulated the problem with a graph cut optimization objective \cite{boykov2001interactive, rother2004grabcut, sormann2006graph}.
% here we talk about the other works that do not use SAM
More recently, several learning-based methods have been proposed for interactive segmentation \cite{ren2022neural, goel2023interactive, yue2023agile3d, kontogianni2023interactive, mirzaei2023spin-nerf, ying2024omniseg3d}. For example, \cite{kontogianni2023interactive} segmented 3D point clouds based on user clicks. Unlike our work, they constructed a dataset for training their model, which limited its utility to parsing objects from a scene.

% these come at the end as they use SAM
Very recently, \cite{kirillov2023segment} presented a foundation model for 2D interactive segmentation termed SAM, which triggered a line of follow-up works aiming at harnessing SAM's impressive capabilities to the 3D domain \cite{cen2023segment, chen2023interactive, yang2023sam3d, 2023arXiv230602245Z}. One approach is to segment 2D projections of the 3D data and fuse them in 3D. However, such an approach requires high user guidance, as the segmentation is performed in 2D, and the user's input is required for different views.

Another approach taken by \cite{chen2023interactive} is to lift SAM's features to a NeRF representation and use SAM's decoder to obtain the segmentation masks. As in the first approach, applying the segmentation in 2D limits the method's capabilities, since it cannot segment together occluded regions in 3D that are not visible concurrently in any 2D view. In contrast to existing works, our model and the user clicks are applied directly in 3D, simplifying the segmentation process and enabling the native segmentation of meaningful regions in 3D (as demonstrated in \cref{fig:view_generalization}).

\section{Method} \label{sec:method}
Given a 3D shape depicted as a mesh $\Mcal$ with vertices $V = \{v_i\}_{i=1}^{n}$, $v_i = (x_i, y_i, z_i) \in \mathbb{R}^3$, and a set of selected vertices by the user representing positive or negative clicks, our goal is to predict the per-vertex probability $P = \{p_i\}_{i=1}^{n}$, $p_i \in [0, 1]$ of belonging to a region adhering to the user inputs. Our system offers an interactive user interface, such that the number of clicks and their type can be varied and the segmented region of the shape is updated accordingly.

We tackle the problem by proposing an interactive segmentation technique consisting of two parts: an encoder that maps vertex coordinates to a deep semantic vector and a decoder that takes the \siga{vertex features} and the user clicks and predicts the corresponding mesh segment. The decoder contains an interactive attention layer supporting a variable number of clicks, which can be positive or negative, to increase or decrease the segmented region. \cref{fig:system} presents an overview of the method.

\subsection{Mesh Feature Field}
\label{sec:encoder}

\siga{Our encoder learns} a function $\phi\colon \mathbb{R}^{3} \rightarrow \mathbb{R}^{d}$ that embeds each mesh vertex into a deep feature vector $\phi(v_i) = f_i$, where $d$ is the feature dimension. \siga{The collection of mesh vertex features is denoted as $F \in \mathbb{R}^{n \times d}$ and regarded as the Mesh Feature Field (MFF).}

The \siga{encoder} distills the semantic information from a pretrained 2D foundation model for image segmentation \cite{kirillov2023segment} and facilitates a 3D consistent feature representation for interactive segmentation of the mesh. To train the \siga{encoder}, we render the high-dimensional vertex attributes differentiably and obtain the 2D projected features:
\begin{equation} \label{eq:rendered_features}
I_f^\theta = \Rcal(\Mcal, f, \theta) \in \mathbb{R}^{w \times h \times d},
\end{equation}

\noindent where $\Rcal$ is a differentiable renderer, $\theta$ is the viewing direction, \siga{$f$ represents the visible vertices in the view,} and $w \times h$ are the spatial dimensions of the rasterized image.

The \siga{encoder} is implemented as a multi-layer perceptron network. To supervise its training, we render the mesh into a color image $I_c^\theta$ and pass it through the encoder $E_{2D}$ of the 2D foundation model~\cite{kirillov2023segment} to obtain a reference feature map:
\begin{equation} \label{eq:ref_features}
I_e^\theta = E_{2D}(I_c^\theta) \in \mathbb{R}^{w \times h \times d}.
\end{equation}

\noindent This process is repeated for multiple random viewing angles $\Theta$, and our encoder is trained to minimize the discrepancy between the rendered MFF and the reference 2D features:
\begin{equation} \label{eq:encoder_loss}
\Lcal_{enc} = \frac{1}{|\Theta|} \sum_{\theta \in \Theta} || I_f^\theta - I_e^\theta ||^2.
\end{equation}

The 2D model operates on each image separately and might produce inconsistent features for different views of shape. In contrast, our MFF is defined in 3D and is view-consistent by construction. It consolidates the information from the multiple views and lifts the 2D embeddings to a coherent field over the mesh surface. Additionally, we emphasize that the MFF is optimized independently of the user inputs. This is a key consideration in our method, resulting in a condition-agnostic representation, describing inherent semantic properties of the mesh. We validate this design choice with an ablation experiment demonstrated in \cref{fig:ablation} and explained in the supplementary. The MFF is optimized until convergence and then utilized together with the user click prompts to compute the interactive mesh partition.

\begin{figure}
\centering
\includegraphics[width=\columnwidth]{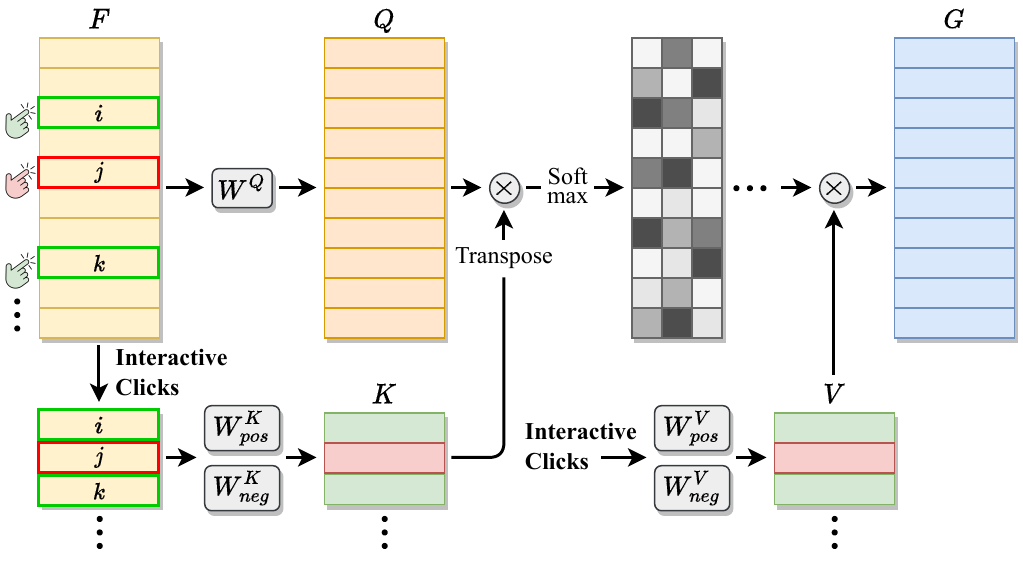}
\caption{\textbf{Interactive Attention.} Our interactive attention layer can handle a variable number of user clicks. The clicks may be positive or negative to indicate region inclusion or exclusion, respectively.
}
\label{fig:interactive_attention}
\end{figure}

\subsection{Interactive Attention Layer}
\label{sec:interactive}

The interactive attention layer is part of the decoding component of our system (\cref{fig:system}). Its structure is illustrated in \cref{fig:interactive_attention}. The layer computes the interaction between the user input clicks and the mesh vertices, accommodating variable numbers and types of clicks, positive and negative. This key element in our method enables a unified decoder architecture supporting various user click settings.

Our interactive attention extends the scaled dot-product attention mechanism \cite{vaswani2017attention}. The features of the positively and negatively clicked vertices are marked as $F_{pos} \in \mathbb{R}^{n_{pos} \times d}$ and $F_{neg} \in \mathbb{R}^{n_{neg} \times d}$, respectively, where $n_{pos}$ and $n_{neg}$ are the number of clicks of each type. The interactive attention layer projects the mesh features $F$ to Queries and the features of the clicked points to Keys and Values:
\begin{equation} \label{eq:qkv}
\begin{split}
Q = & \: F W^Q \in \mathbb{R}^{n \times d} \\
K_{\{pos, neg\}} = & \: F_{\{pos, neg\}} W^K_{\{pos, neg\}} \in \mathbb{R}^{n_{\{pos, neg\}} \times d} \\
V_{\{pos, neg\}} = & \: F_{\{pos, neg\}} W^V_{\{pos, neg\}} \in \mathbb{R}^{n_{\{pos, neg\}} \times d},
\end{split}
\end{equation}

\noindent where $W^Q, W^K_{\{pos, neg\}}, W^V_{\{pos, neg\}} \in \mathbb{R}^{d \times d}$ are learnable weight matrices. Then, the mesh vertices are attended to the user clicks to obtain the conditioned mesh features:

\begin{equation} \label{eq:attention}
G = \text{softmax}\left(\frac{QK^T}{\sqrt{d}}\right)V \in \mathbb{R}^{n \times d},
\end{equation}

\noindent where $K, V \in \mathbb{R}^{(n_{pos} + n_{neg}) \times d}$ are the concatenation of $K_{pos}, K_{neg}$ and $V_{pos}, V_{neg}$, respectively.

Our attention mechanism condenses variable interactive user inputs into a fixed-length output. It learns the representation of positive and negative clicks, correlates the mesh vertices with them, and yields updated vertex features to enable the on-the-fly segmentation of the shape. Another benefit of our attention layer is that it is permutation invariant \wrt the user clicks. In other words, it is independent of the sequential order of the point clicks and consistent in their joint influence on the shape partition. Moreover, the attention's output $G$ is permutation equivariant \wrt the vertex order in $F$, a desired property for the mesh data structure. 

%\ifoverpic
%\input{figures/view_generalization/view_generalization_png.tex}
%\else
\begin{figure}
\centering
\newcommand{\pl}{-2}
% trim: left, bottom, right, top
\includegraphics[width=\linewidth, trim=190 0 210 20, clip]{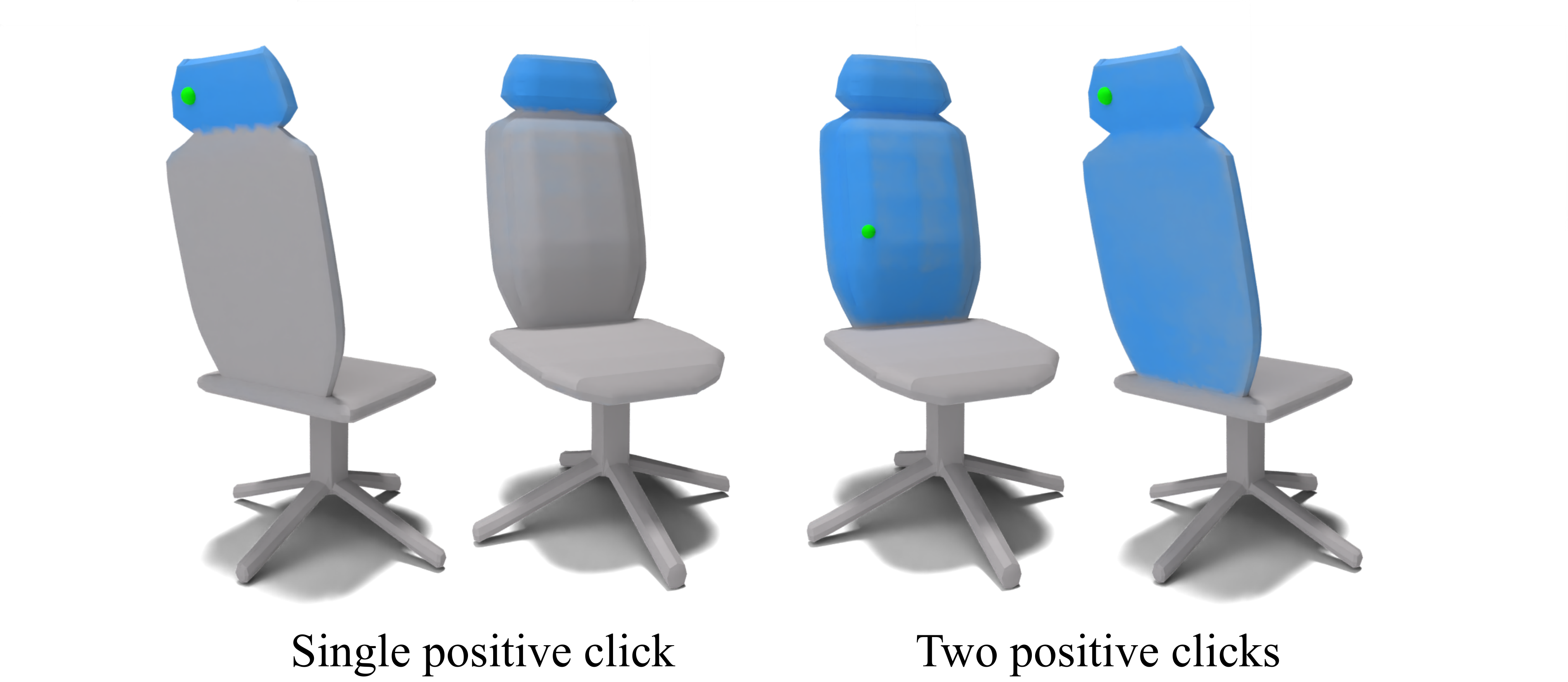}
\vspace{-5mm}
\caption{\textbf{Native 3D segmentation.} \ourmethod{} segments parts in a 3D-consistent manner, regardless of whether the surface is occluded from the point click. A point is selected on the back of the chair (left), which is not visible from the front view. Still, our method delineates the occluded surface even though the 2D training data cannot contain this information. Furthermore, we may input two point clicks occluded from each other, one on the back of the chair and one on the front (right). These points cannot be simultaneously input to any 2D decoder, as they are not visible concurrently from any single viewpoint. Nonetheless, \ourmethod{} faithfully segments the whole backrest part.}
\label{fig:view_generalization}
\end{figure}

%\fi

\begin{figure*}
\centering
\newcommand{\pl}{-2}
% trim: left, bottom, right, top
\includegraphics[width=\linewidth, trim=110 0 140 0, clip]
{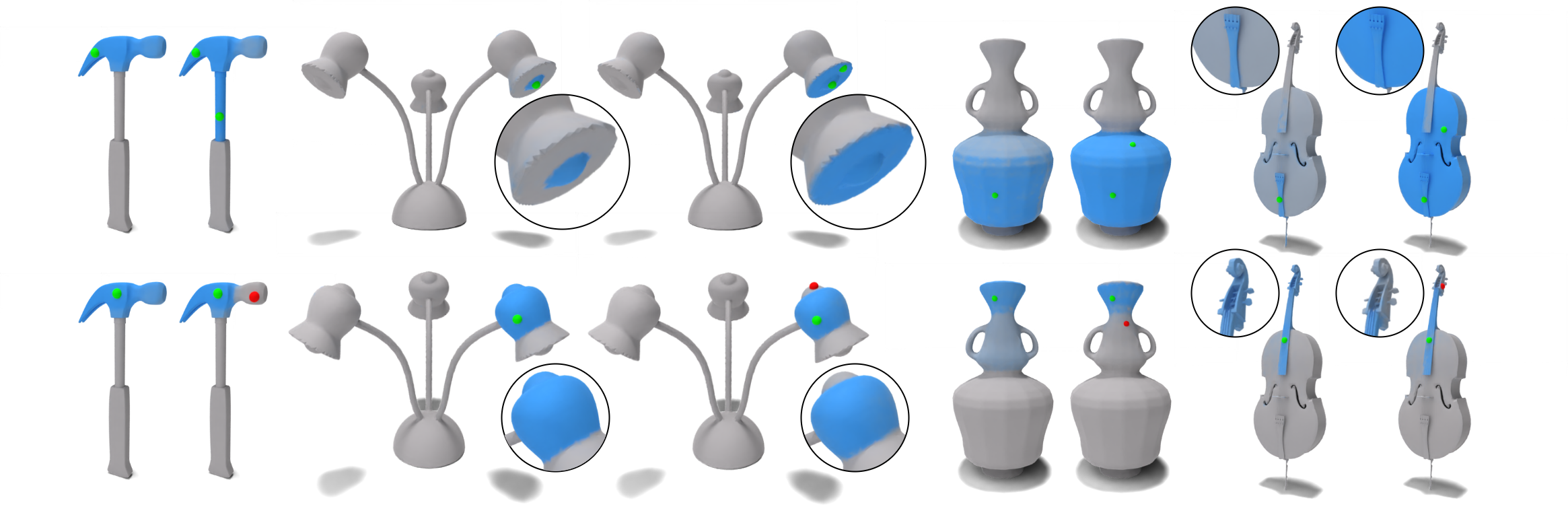}
\vspace{-5mm}
\caption{\textbf{Couple of clicks results.} \ourmethod{} produces fine-grained segmentations from a couple of clicks (both positive and negative) as input. Each pair of shapes starts with a single positive click (left), which can be further customized using an additional click (right).}
\label{fig:multi}
\end{figure*}

\subsection{Segmentation Prediction}
\label{sec:decoder}

The output of our model is a segmentation of the mesh that adheres to the user clicks, represented as the per-vertex probability of belonging to the desired region. To do so, we learn to decode the a posteriori condition dependent vertex features $g_i = [G]_i$ and the a priori inherent embedding $f_i = [F]_i$ into the partition probability:
\begin{equation} \label{eq:seg_prob}
p_i = \psi(f_i, g_i) \in [0, 1].
\end{equation}

\noindent $\psi$ is implemented as a multi-layer perceptron network, where $f_i$ and $g_i$ are concatenated along the feature dimension at the network's input. The remaining question is -- how to supervise the training of such an obscure and ill-defined problem?

Similar to our encoder's training, we translate the problem to the 2D domain and harness the power of the 2D foundation model \cite{kirillov2023segment} for our 3D decoder learning (\cref{fig:system}). We project the mesh probability map with a differentiable rasterizer to a probability image $I_p^{\theta'} = \Rcal(\Mcal, p, \theta') \in [0, 1]^{w' \times h' \times 2}$, where $\theta'$ is the viewing angle, and the image channels represent the segment and background probabilities. Then, we project the 3D clicks to their corresponding 2D pixels and use them as prompts to segment the rendered color mesh image with the 2D segmentation model, resulting in the supervising probability mask $I_m^{\theta'} \in \{0, 1\}^{w' \times h' \times 2}$. We randomize the viewing direction $\theta'$ and train our decoder subject to the optimization objective:
\begin{equation} \label{eq:dec_loss}
\Lcal_{dec} = \frac{1}{|\Theta'|} \sum_{\theta' \in \Theta'} \text{CE} (I_p^{\theta'}, I_m^{\theta'}),
\end{equation}

\noindent where $\text{CE}$ denotes the binary cross-entropy loss. 

To prepare data for our 3D decoder training, we simulate user clicks and generate masks from the 2D model \cite{kirillov2023segment}. The data generation process includes two phases. First, we pick a \camrdy{small training} subset of \camrdy{3\% of the} mesh vertices \siga{well distributed over the shape using Farthest Point Sampling \cite{eldar1997FPS}}, where each vertex is regarded as a single positive click. For each vertex, we generate random views, feed each one through the 2D foundation model, and get the reference segmentation mask. Then, for each view, we sample another \camrdy{training} vertex visible within the viewing angle, which is set to be a positive or a negative click, and compute the updated segmentation by the 2D model. According to its type, we require the second click to increase or decrease the previous segmentation mask to obtain rich and diverse training data. 

As seen in \cref{fig:system}, the supervision signal is highly inconsistent. The 3D shape and the same clicks are interpreted differently by the 2D model, yielding strong variations in the 2D segmentation masks. Nevertheless, our method reveals a coherent underlying 3D segmentation function out of the noisy 2D measurements. Our decoder utilizes the robust distilled 3D vertex features, applies their interaction with the clicked points, and computes the region probability map directly in 3D. \ourmethod{} segmentaitons are view-consistent \textit{by construction}, improving substantially over its training data. Furthermore, although trained with only 2D supervision, \ourmethod{} delineates meaningful regions in 3D that are not entirely visible in a single 2D projection (\cref{fig:view_generalization}).

\section{Experiments} \label{sec:experiments}

We evaluate \ourmethod{} in a variety of aspects. First, we demonstrate the generality and fidelity of our method in~\cref{sec:generality,sec:fidelity}, respectively. Then, in~\cref{sec:properties}, we showcase \siga{the generic feature information captured by \ourmethod{}}. Finally,~\cref{sec:generalization} presents the strong generalization power of \ourmethod{} in terms of the selected point, views of the click, and the number of clicks.

We apply our method to diverse meshes from different sources: COSEG~\cite{coseg_2011}, Turbo Squid~\cite{turbosquid}, Thingi10K~\cite{Thingi10K}, Toys4k~\cite{Toys4k}, SCAPE \cite{anguelov2005scape}, SHREC '19 \cite{melzi2019shrec}, ModelNet~\cite{modelnet}, ShapeNet~\cite{chang2015shapenet}, \siga{and PartNet~\cite{mo2019partnet}}. \ourmethod{} is highly robust to the shape properties. It operates on meshes with different numbers of vertices and various geometries, including thin, flat, and high-curvature surfaces.

\ourmethod{} is implemented in PyTorch~\cite{pytorch} and its training time varies according to the number of mesh vertices. For a mesh with $3000$ vertices, the optimization takes about 3 hours on a single Nvidia A40 GPU. \siga{Training the model is a one-time offline phase. Once trained,} querying the model with input clicks takes only about 0.7 seconds\siga{, which allows fast interactive interaction with the shape}. In our experiments, we used SAM ViT-H with an image size of $224 \times 224$. To obtain fine-grained segmentations, we utilized the smallest scale mask from SAM for the projected clicked points. Additional details are provided in the supplementary.

%%%%%%%%%%%%%%
% Generality %
%%%%%%%%%%%%%%
\subsection{Generality of \ourmethod{}} \label{sec:generality}
\ourmethod{} is highly versatile and works on a variety of shapes and geometries. It is not limited to any specific shape category nor a pre-defined set of parts and can be applied to meshes from various domains, including humanoids, animals, musical instruments, household objects, and more. Our method is also applicable to shapes with complex geometric structures and is optimized to capture the elements of the given object.

\cref{fig:single} presents different single-click results. \ourmethod{} successfully segments regions with sharp edges, such as the neck of the lamp and the thin spokes of the bike. It also captures accurately the flat surface of the alien's head and the curved lower part of its leg. Moreover, \ourmethod{} can segment small parts of the shape - the bike's seat and the water bottle, or larger portions, such as the body parts of the camel.

%%% quantitative results table %%%
\begin{table}[t!]
\caption{\siga{\textbf{Quantitative evaluation on PartNet.} We compare the segmentation performance of different clicked-based interactive techniques on shapes from the PartNet dataset \cite{mo2019partnet}. IoU stands for Intersection over Union. \ourmethod{}'s scores are substantially higher than those of the alternatives.
}}
\vspace{-5pt}
\centering
\begin{tabular}{lccc}
\toprule
Method  &  InterObject3D & SAM Baseline & \ourmethod{} (ours) \\
\midrule
Accuracy $\uparrow$ & 0.54  & 0.76 & \textbf{0.95}  \\
IoU $\uparrow$ & 0.38 & 0.51 & \textbf{0.90}  \\
\bottomrule
\end{tabular}
\vspace{-5pt}
\label{tab:quantitative_resutls}
\end{table}

%%%%%%%%%%%%%%%%%
%%% Fidelity %%%
%%%%%%%%%%%%%%%%%
%% how well iSeg adheres to the click (single and multi-click gallery)
\subsection{Fidelity of \ourmethod{}} \label{sec:fidelity}
Our method's training is supervised by segmentation masks generated from SAM~\cite{kirillov2023segment} for 2D renderings of the shape. As we show in \cref{fig:system}, SAM's masks differ substantially between views. In contrast, \ourmethod{} manages to fuse the noisy training examples into a coherent 3D segmentation model that corresponds to the clicked vertices. Examples are presented in \cref{fig:single,fig:multi}. In the supplementary, we further demonstrate the method's 3D  consistency.

\ourmethod{} is adapted to the granularity of the given mesh, which enables it to adhere to the user's clicks and segment the region of the shape related to the user's inputs. For example, in \cref{fig:single}, for a click on the alien's middle antenna, the entire antenna and \textit{only that particular antenna} is selected. We see similar behavior for other clicks, such as the one on the lamp's base and the camel's neck.

In \cref{fig:multi}, we further show the results of \ourmethod{} for a couple of clicks. We incorporate either a second positive click that extends the segmented part or a second negative click that retracts the region. For example, with the first click, the fine-grained bulb area of the lamp is segmented. Then, the second positive click enables to include the flat surface surrounding the bulb. The negative click, on the other hand, offers the control to reduce and refine the segmentation region. As shown in \cref{fig:multi}, the first click on the hammer's head segments the entire head. The front part can be easily and intuitively removed by a second negative click on it.

\siga{
%\smallskip
%\noindent \textbf{Quantitative evaluation.}
%\subsubsection{Quantitative evaluation}
\paragraph{Quantitative evaluation}
As far as we can ascertain, there are no annotated datasets for click-based interactive segmentation of 3D shapes. Thus, we adapted the part segmentation dataset PartNet \cite{mo2019partnet} for our setting. The evaluation included 170 meshes sourced from all the categories in the dataset. For each mesh, we selected five test vertices from a part at random, where each vertex was regarded as a single click, and measured how well the part was segmented. We used two evaluation metrics: Accuracy (ACC) and Intersection over Union (IoU). Further details about the evaluation are provided in the supplemental material.
}

We considered two alternatives for comparison, InterObject3D \cite{kontogianni2023interactive} and a baseline we constructed based on SAM's 2D segmentations. InterObject3D is a recent work on interactive segmentation of 3D objects. \camrdy{In our experiments, we employed the publicly available pretrained model released by the authors.}

For the SAM baseline, we rendered the shape and projected the clicked point to 2D from 100 random views, computed SAM's mask, and re-projected the result back to 3D for each visible vertex. Then, we averaged the predictions according to the number of times each vertex was seen. \siga{In the supplementary material, we discuss additional baselines we devised using SAM.} Other interactive segmentation techniques use an implicit 3D representation \cite{chen2023interactive} or perform a different task (object detection) \cite{2023arXiv230602245Z}, and thus, are not directly comparable to our method.

\begin{figure}[!t]
\centering
\newcommand{\pl}{-0.5}
% trim: left, bottom, right, top
\includegraphics[width=\linewidth, trim=0 0 0 60, clip]{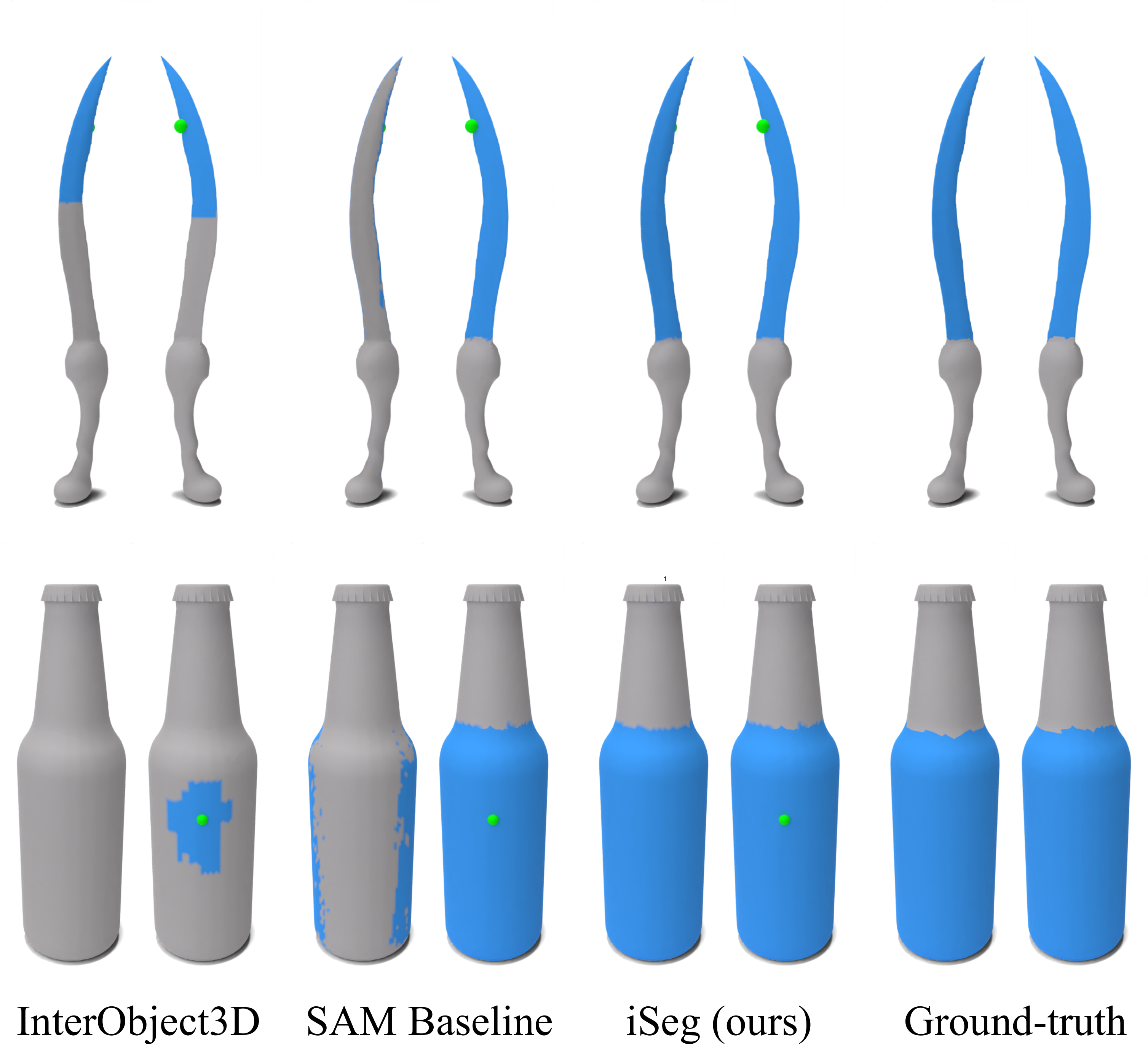}
\vspace{-7mm}
\caption{\siga{\textbf{Segmentation comparison on PartNet.} Each pair shows different views of the segmentation result for different interactive techniques. The other methods produce a partial segmentation of the part containing the click. In contrast, \ourmethod{} obtains an accurate result, which is similar to the ground-truth annotation of the part.}}
\label{fig:partnet}
\end{figure}

\siga{
\cref{tab:quantitative_resutls} presents the ACC and IoU averaged over the test clicks and shapes, and \cref{fig:partnet} shows visual examples. InterObject3D does not select the part properly and yields a partial segmentation. SAM Baseline segments the region of the part where the click is visible. However, occluded regions that belong to the part are not marked. In contrast, \ourmethod{} adheres to the clicked point. It segments a coherent region in 3D, which is similar to the ground-truth part label, and achieves much higher ACC and IoU than the baselines.
}

%\smallskip
%\noindent \textbf{Perceptual user study.}
%\subsubsection{Perceptual user study}
\paragraph{Perceptual user study}
\siga{\ourmethod{} is not limited to a particular shape type from a dataset or specific parts defined in the dataset. In such cases, ground-truth labels are unavailable. Thus, to evaluate the effectiveness of the flexible and diverse segmentations offered by our method}, we opt to perform a perceptual user study. We used 20 meshes from different categories, such as humanoids, animals, and man-made objects, and included 40 participants in our study.

%%% perceptual study table %%%
\begin{table}[t!]
\caption{\textbf{Perceptual user study.} We evaluate the 3D segmentation effectiveness on a scale of 1 to 5, corresponding to completely ineffective and completely effective segmentation. Our method is considered much more effective than the competitors.}
\vspace{-5pt}
\centering
\begin{tabular}{lccc}
\toprule
Method  &  InterObject3D & SAM Baseline & \ourmethod{} (ours) \\
\midrule
Effectiveness $\uparrow$ &  2.54 & 3.02 & \textbf{4.55}  \\
\bottomrule
\end{tabular}
\vspace{-2pt}
\label{tab:perceptual_study}
\end{table}

For each mesh, we showed the 3D segmentation for a clicked point from multiple viewing angles and asked the participants to rate the effectiveness of the result on a scale of 1 to 5. The score 5 refers to a completely effective segmentation, where the entire 3D region corresponding to the clicked point is selected. When part of the 3D region is marked, the segmentation is considered partially effective. The score 1 is defined as a completely ineffective segmentation and refers to no region selection. Examples are presented in \cref{fig:effectiveness}.

\cref{tab:perceptual_study} summarizes the effectiveness score averaged over all the meshes and participants. \cref{fig:sam_occlude} further compares our method with the SAM baseline. As seen in the figure and reflected by the table, the participants rated the effectiveness of \ourmethod{} much higher than the other methods, indicating its fidelity to the clicked point.

\siga{
%\smallskip
%\noindent \textbf{Applications.}
%\subsubsection{Applications}
\paragraph{Applications}
\ourmethod{} computes contiguous and localized shape partitions. These segmentations can be used to extract shape parts easily and enable applications such as full-shape segmentation and local geometric editing of the mesh. These results are demonstrated in \cref{fig:separability_supp,fig:full_seg_supp,fig:edits_supp}. Further details and discussion appear in the supplementary material.
}

%%%%%%%%%%%%%%%%%%%%%%%%%%%%%%%
% Generic Feature Information %
%%%%%%%%%%%%%%%%%%%%%%%%%%%%%%%
\subsection{Generic Feature Information} \label{sec:properties}

%%% PartNet figure %%%
%\ifoverpic
%\input{figures/partnet/partnet_png.tex}
%\else

%\fi

\begin{figure}
\centering
\newcommand{\pl}{-2}
% trim: left, bottom, right, top
\includegraphics[width=\linewidth, trim=210 0 130 0, clip]{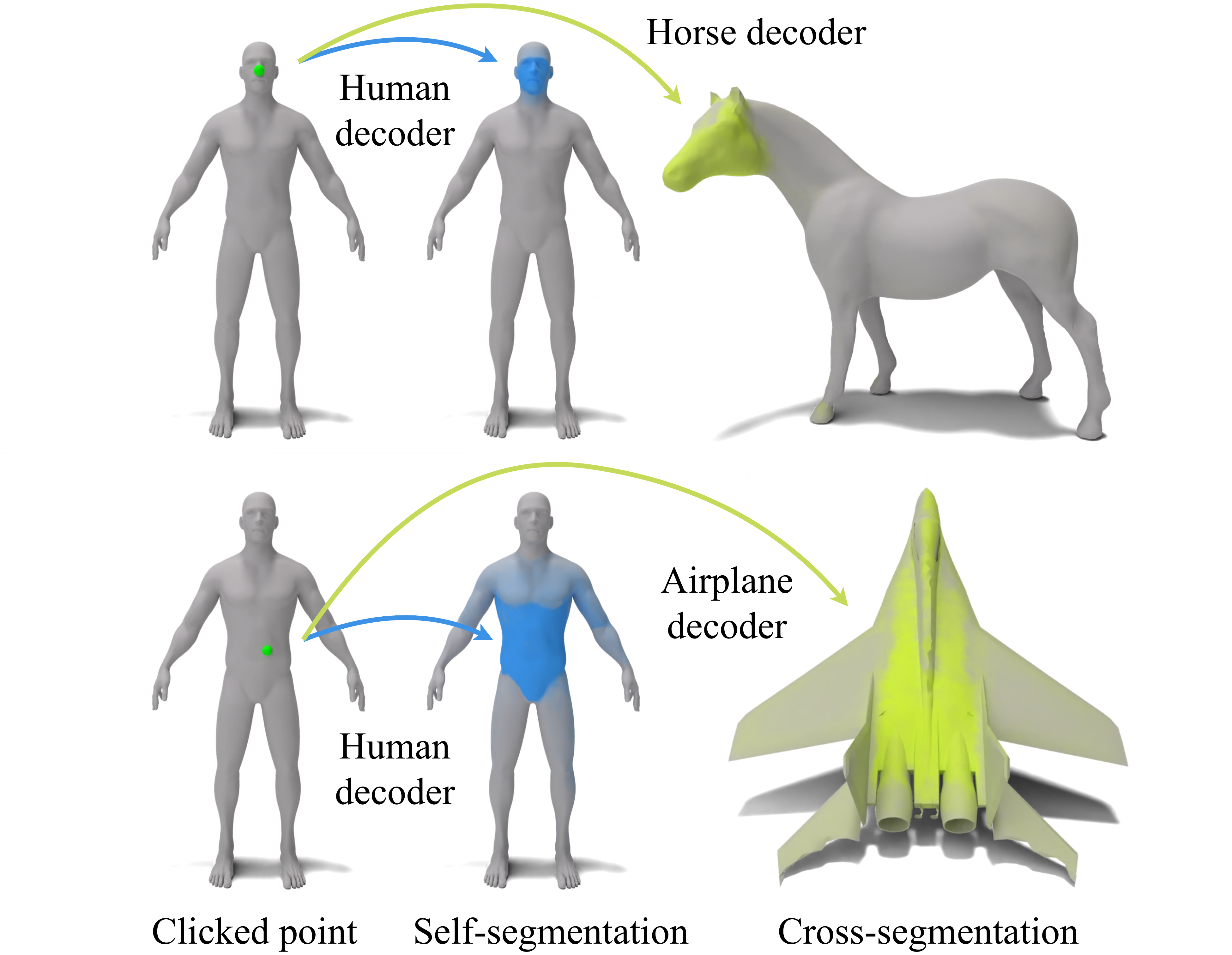}
\vspace{-6mm}
\caption{\textbf{Cross-domain segmentation.} \ourmethod{} optimizes a condition-agnostic feature field, which is capable of transferring between shapes. The feature vector of a point click of one mesh (left) is used to segment the same shape (middle) as well as \textit{another} shape from a \textit{different} domain (right).}
\label{fig:cross_seg}
\end{figure}

Our mesh feature field is distilled directly from SAM's encoder and is independent of the user's clicks for segmentation. Thus, although \ourmethod{} is optimized per mesh, the semantic feature representation is shared across shapes. We demonstrate this property by cross-domain segmentation. In this experiment, we use the point encoded features from one shape and predict region probability with a \textit{different} shape's decoder. \cref{fig:cross_seg} shows examples. The transferable features enable the creation of cross-domain shape analogies, such as how the belly of a human corresponds to the \textit{``belly''} of an airplane.

%%%%%%%%%%%%%%%%%%
% Generalization %
%%%%%%%%%%%%%%%%%%
% vertex generalization
% view generalization
% number of clicks generalization
\subsection{Generalization Capabilities} \label{sec:generalization}

%\noindent \textbf{Unseen \camrdy{mesh} vertices.}
%\subsubsection{Unseen \camrdy{mesh} vertices}
\paragraph{Unseen \camrdy{mesh} vertices}
Our method exhibits strong generalization power. First, we emphasize that we train just on a small fraction of 3\% of the mesh vertices. Still, \ourmethod{} is successfully applied to other \camrdy{mesh} vertices unseen during training and properly respects the clicked points, as shown in \cref{fig:single,fig:multi} and discussed in \cref{sec:fidelity}. We note that all the results shown in the paper are for test vertices.

%%% cross-segmentaiton figure %%%
%\ifoverpic
%\input{figures/cross/cross_png.tex}
%\else

%\fi

%\smallskip
%\noindent \textbf{Unseen views.}
%\subsubsection{Unseen views}
\paragraph{Unseen views}
Although \ourmethod{} was trained with 2D supervision only, its predictions are 3D in nature. \cref{fig:view_generalization} exemplifies this phenomenon. A click on the back side of the backrest segments the front side as well. Such supervision does not exist for our model's training, since the clicked vertex cannot be seen from the front side. Similarly, two clicks at opposite sides of the backrest segment the entire part, although they cannot be seen together from \textit{any single 2D view}. This result suggests that \ourmethod{} learned 3D-consistent semantic vertex information, enabling it to generalize beyond its 2D supervision. 

\begin{figure}[!t]
\centering
% trim: left, bottom, right, top
\includegraphics[width=0.99\linewidth, trim=0 0 0 -44.5]{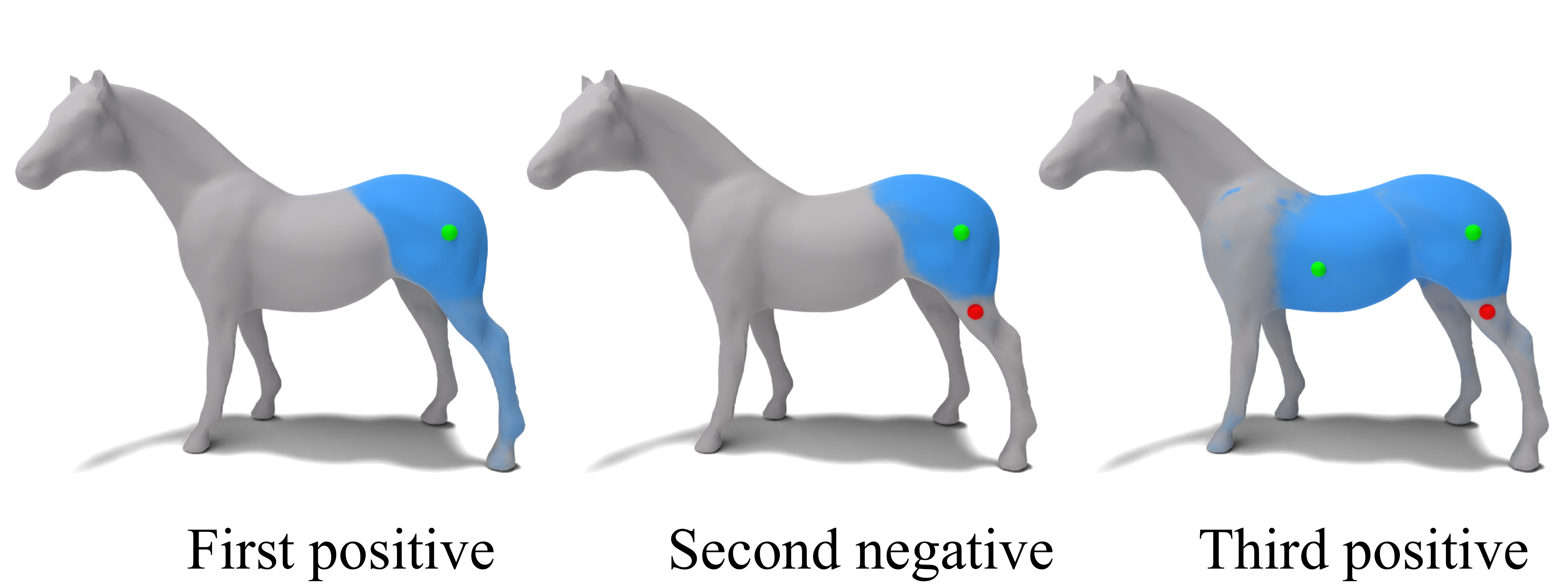}
\vspace{-1mm}
\captionof{figure}{\textbf{Customized segmentations.} \ourmethod{} is capable of creating customized segmentations specified by several input clicks.}
\label{fig:sequential_seg}
\end{figure}

%\smallskip
%\noindent \textbf{Unseen number of clicks.}
%\subsubsection{Unseen number of clicks}
\paragraph{Unseen number of clicks}
For resource efficient training, we trained \ourmethod{} on up to two clicks: single click, second positive click, and second negative click. Nonetheless, our model offers customized segmentation with more than two clicks, as demonstrated in \cref{fig:sequential_seg}. We attribute this capability to the interactive attention mechanism. The interactive attention layer seemed to learn the representation of a positive and a negative click, and how to attend to each click for a meaningful multi-click segmentation.

\begin{figure}[!t]
\centering
\newcommand{\pl}{-2}
% trim: left, bottom, right, top
\includegraphics[width=0.8\linewidth, trim=30 0 30 100, clip]{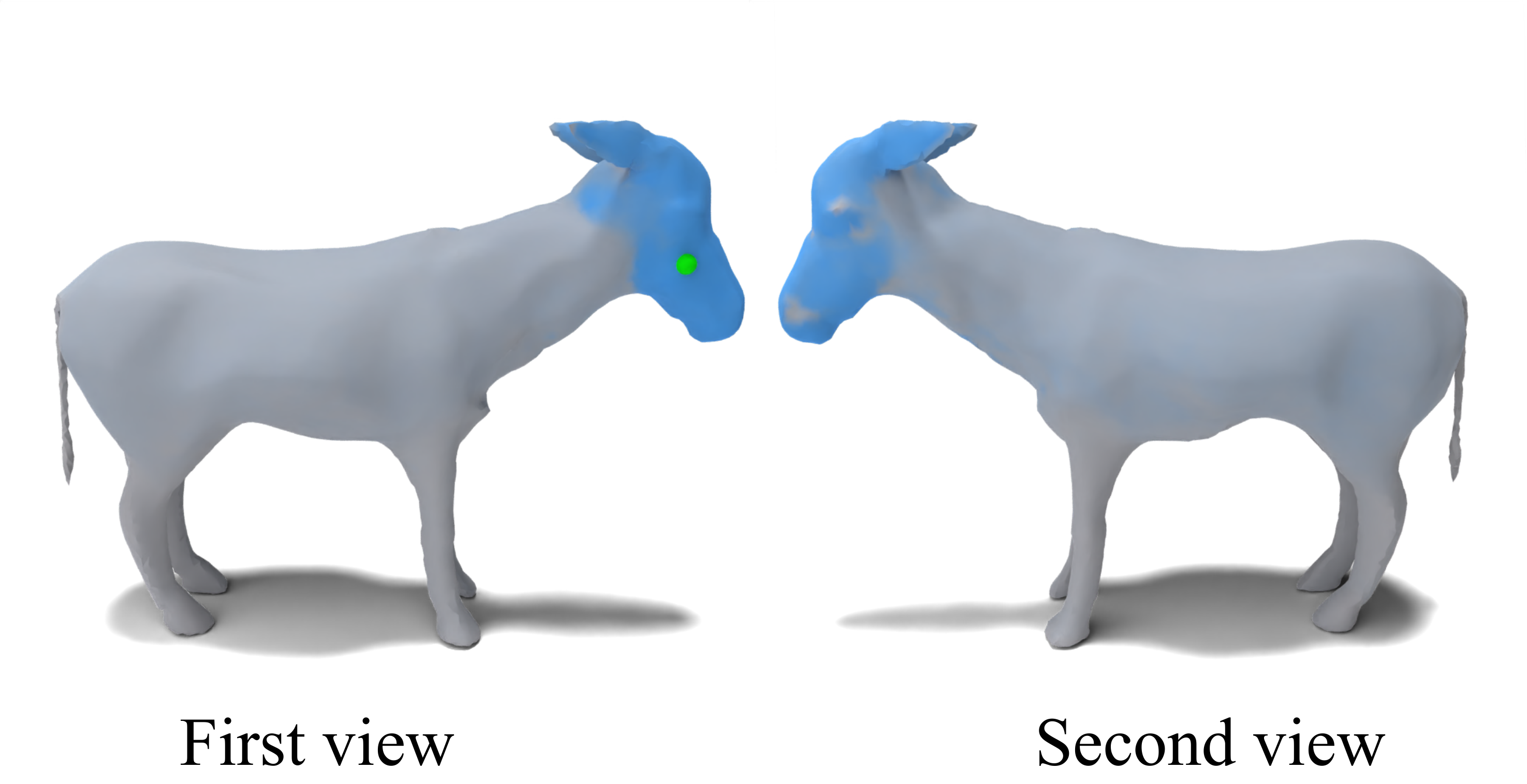}
\vspace{-2mm}
\captionof{figure}{\textbf{Limitation.} \ourmethod{} may not produce a symmetric segmentation result for a symmetric shape. In this case, the segmented region of one side of the shape is different than the other.}
\label{fig:limitation}
\end{figure}

\siga{
%\smallskip
%\noindent \textbf{Limitations.}
%\subsubsection{Limitations}
\paragraph{Limitations}
Our method may not follow the symmetry of the mesh exactly, as exemplified in \cref{fig:limitation}. For a click on the goat's head, the segmented regions from the sides of the head differ somewhat from each other, since \ourmethod{} is not trained to segment those regions the same. \camrdy{In the supplementary, we discuss the potential limitation of processing 3D shapes using MLPs operating in the Euclidean space.}
}

\section{Conclusion} \label{sec:conclusion}

In this work, we presented \ourmethod{}, a technique for interactively generating fine-grained tailored segmentations of 3D meshes. We opt to lift features from a powerful pre-trained 2D segmentation model onto a 3D mesh, which can be used to create customized user-specified segmentations. Our mesh feature field is general and may be used for additional tasks such as cross-segmentation across shapes of different categories (\eg~\cref{fig:cross_seg}). Key to our method is an interactive attention mechanism that learns a unified representation for a varied number of positive or negative point clicks. Our 3D-consistent segmentation enables selecting points across occluded surfaces and segmenting meaningful regions directly in 3D (\eg~\cref{fig:view_generalization}).

In the future, we are interested in exploring additional applications of \ourmethod{} beyond segmentation. We have demonstrated that it can potentially be used for cross-domain segmentation, and there may be other exciting applications, such as key-point correspondence, texture transfer, and more.

\begin{acks} \label{sec:acknowledgements}
This research was supported by grant \#2022363 from the United States - Israel Binational Science Foundation (BSF), grant \#2304481 from the National Science Foundation (NSF), and gifts from Adobe, Snap, and Google. We thank the University of Chicago and the Toyota Technological Institute at Chicago (TTIC) for allocating computational resources for this work, and their technical staff for the support along the project. We also extend our gratitude to the members of the 3DL lab at the University of Chicago for their helpful feedback and excellent advice.
\end{acks}

%%%%%%%%% REFERENCES
\bibliographystyle{ACM-Reference-Format}
\bibliography{references.bib}

% first page
\begin{figure*}[!t]
\centering
% trim: left, bottom, right, top
\includegraphics[width=0.99\linewidth, trim=0 0 0 10, clip]
{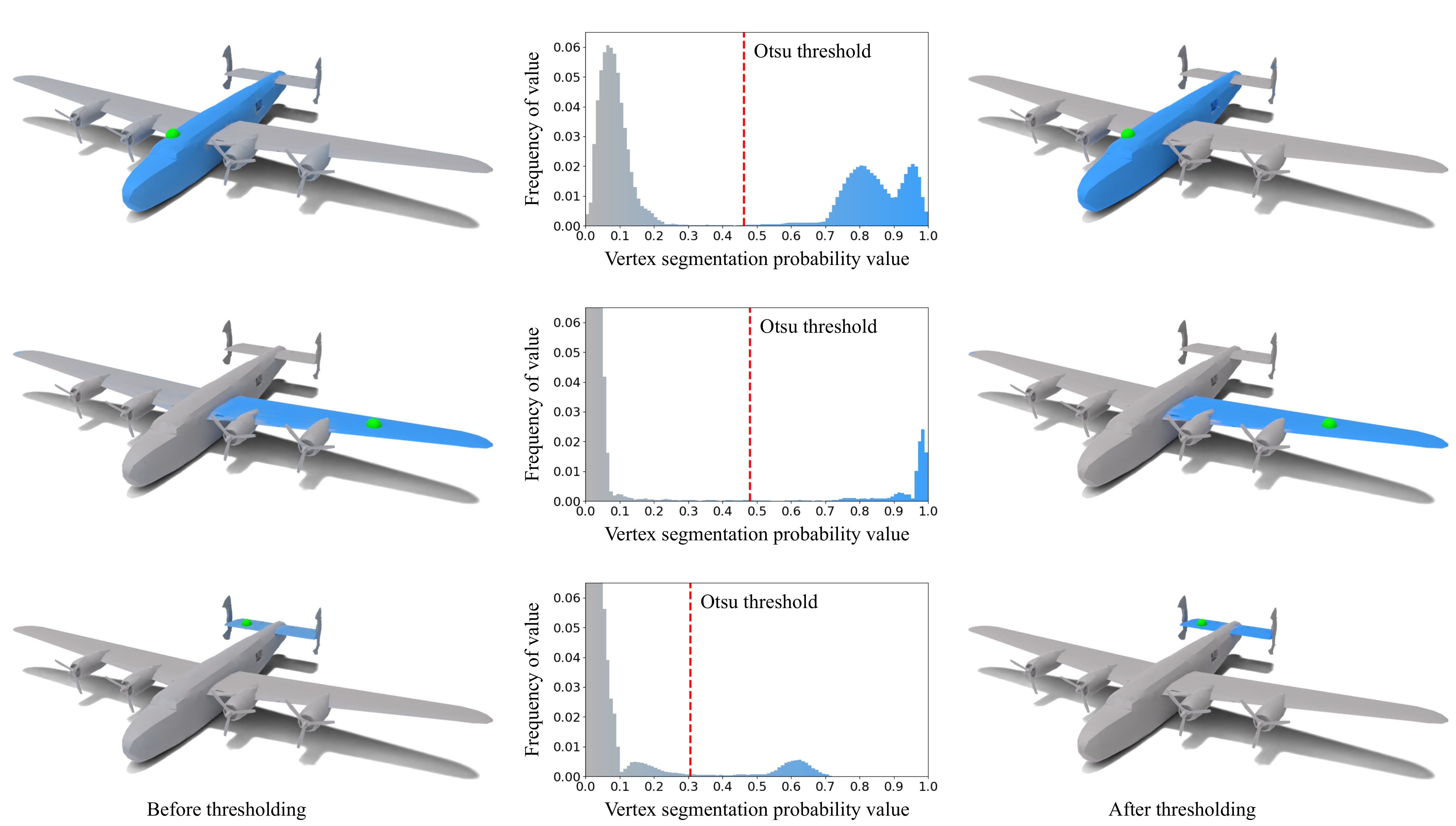}
\caption{\textbf{Segmentation separability.} The per-vertex segmentation probability can be separated distinctively into low and high value populations, which enables the hard selection of the segmented part by simple thresholding, namely, the Otsu threshold \cite{ostu1979threshold}.}
\label{fig:separability_supp}
\end{figure*}

\begin{figure*}[!t]
\centering
% trim: left, bottom, right, top
\includegraphics[width=\linewidth, trim=120 100 0 0, clip]
{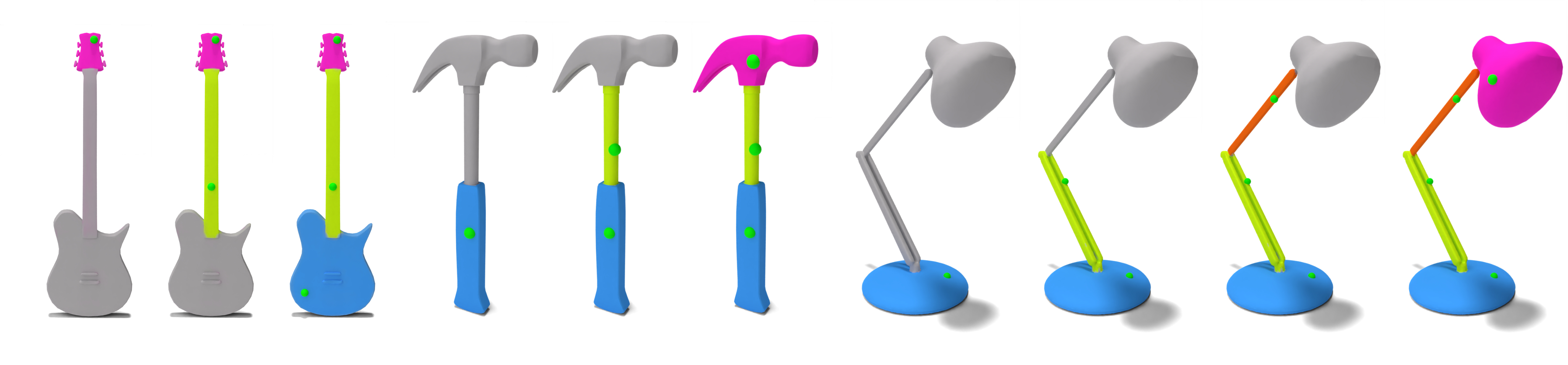}
\caption{\siga{\textbf{Full-shape segmentation.} \ourmethod{} can be used to segment the entire shape with a sequence of interactive clicks. We visualize the segmented parts with different colors and the background region is shown in gray.}}
\label{fig:full_seg_supp}
\end{figure*}

%\ifoverpic
%\input{figures/sequential/sequential_png.tex}
%\else

%\fi

%\ifoverpic
%\input{figures/limitation/limitation_png.tex}
%\else

%\fi

% second page
%\ifoverpic
%\input{figures/edits/edits_png.tex}
%\else
\begin{figure*}[tb!]
\centering
% trim: left, bottom, right, top
\includegraphics[width=\linewidth, trim=150 170 210 0, clip]{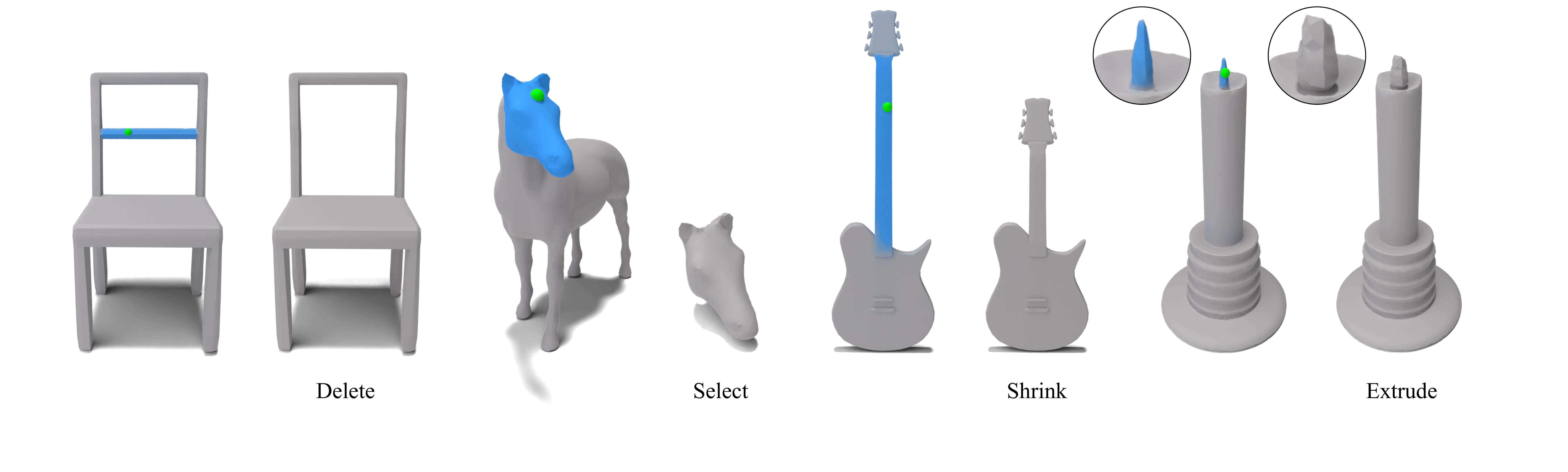}
\caption{\textbf{Local geometric edits.} Our localized and contiguous segmentations enable various shape edits, such as deleting or selecting the segmented region, shrinking it, or extruding it along the surface normal.}
\label{fig:edits_supp}
\end{figure*}

%\fi

%\ifoverpic
%\input{figures/effectiveness/effectiveness_png.tex}
%\else
\begin{figure*}
\begin{center}
\centering
% trim: left, bottom, right, top
\includegraphics[width=\textwidth, trim=0 0 0 0, clip]{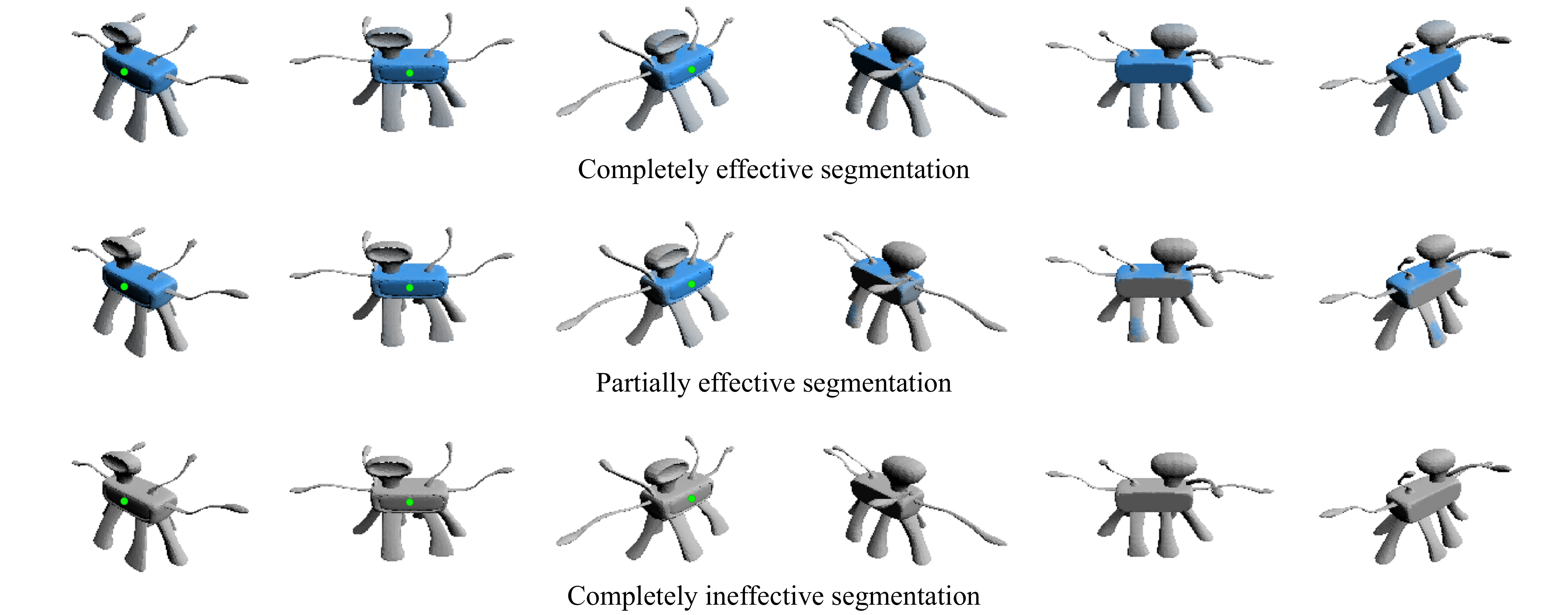}
%\vspace{0.5mm}
\captionof{figure}{\textbf{Segmentation effectiveness.} We visualize results with a varying level of effectiveness, as presented in our perceptual user study. The segmentations from top to bottom rows are considered completely effective, partially effective, and completely ineffective, respectively.}
    \label{fig:effectiveness}
\end{center}
\end{figure*}

%\fi

%\ifoverpic
%\input{figures/occluded/occluded_png.tex}
%\else
\begin{figure}[!tbh]
\begin{center}
\centering
% trim: left, bottom, right, top
\includegraphics[width=0.99\linewidth, trim=0 0 0 0, clip]
{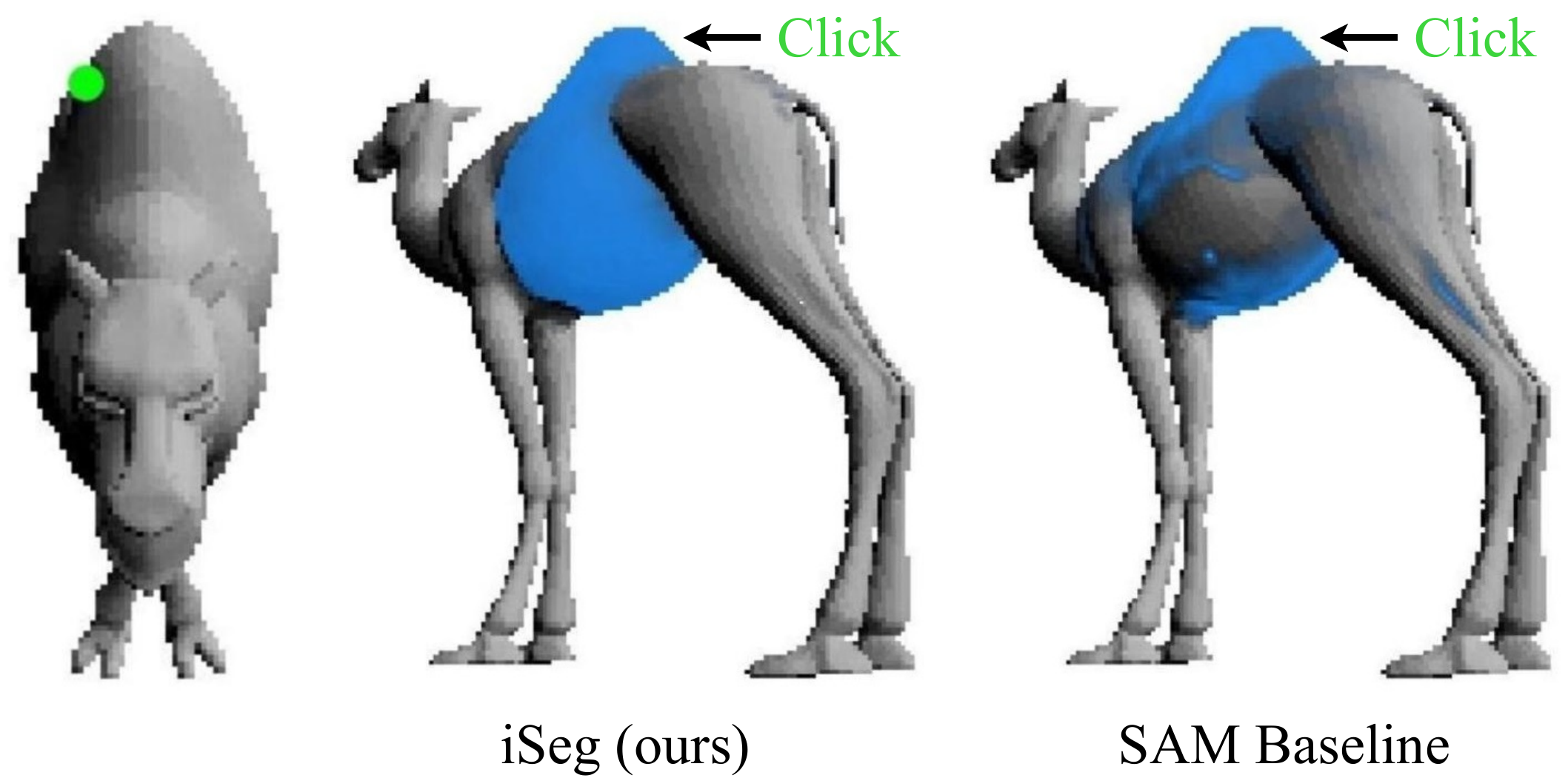}
\vspace{-1mm}
\caption{\textbf{The power of \ourmethod{} for occluded point click.} When the point click (green dot in the leftmost image) is occluded, \ourmethod{} can produce an effective 3D segmentation (highlighted in blue), whereas the SAM baseline is unable to do so.}
\label{fig:sam_occlude}
\end{center}
\end{figure}

%\fi

%\ifoverpic
%\input{figures/ablation/ablation_jpg.tex}
%\else
\begin{figure}[!tbh]
\centering
% trim: left, bottom, right, top
\includegraphics[width=0.99\linewidth, trim=0 0 0 -120, clip]
{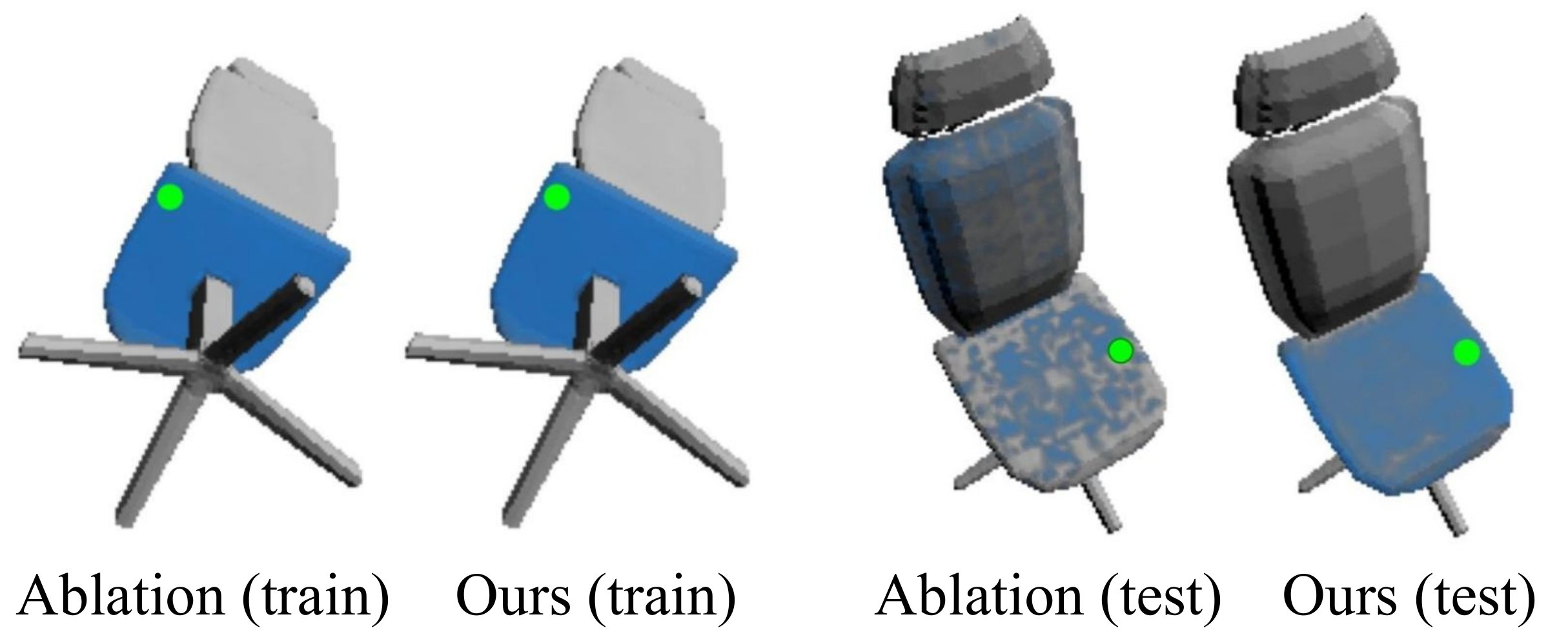}
\vspace{-1.5mm}
\captionof{figure}{\textbf{Ablation test.} We compare an \ourmethod{} model from separate training of the encoder and decoder with an ablation model from joint training of both components. Our proposed separate training scheme results in better generalization for test vertices.}
\label{fig:ablation}
\end{figure}

%\fi

\clearpage

% WARNING: do not forget to delete the supplementary pages from the submission 
\ifwithsupp
\clearpage
\appendix
\maketitlesupplementary

% How to split pdf file:
%To split the supplementary pages from the main paper, you can use \href{https://support.apple.com/en-ca/guide/preview/prvw11793/mac#:~:text=Delete%20a%20page%20from%20a,or%20choose%20Edit%20%3E%20Delete).}{Preview (on macOS)}, \href{https://www.adobe.com/acrobat/how-to/delete-pages-from-pdf.html#:~:text=Choose%20%E2%80%9CTools%E2%80%9D%20%3E%20%E2%80%9COrganize,or%20pages%20from%20the%20file.}{Adobe Acrobat} (on all OSs), as well as \href{https://superuser.com/questions/517986/is-it-possible-to-delete-some-pages-of-a-pdf-document}{command line tools}.

The following sections provide more information regarding our interactive segmentation method. Appendix \ref{sec:additional_results} shows additional results and experiments we conducted with \ourmethod{}. Appendices \siga{\ref{sec:quantitative_evaluation} and} \ref{sec:user_study} discuss our \siga{quantitative evaluation on the PartNet dataset and the} perceptual user study, respectively. In appendix \ref{sec:ablation_tests}, we detail the results of an ablation test. Finally, in appendix \ref{sec:implementation_det}, we elaborate on our implementation details for the encoder and decoder networks and the comparison settings to previous work.

\section{Additional Results}
\label{sec:additional_results}

\siga{
%\noindent \textbf{Fidelity.}
\paragraph{Fidelity}
\ourmethod{} is flexible and not bounded to parts defined in a dataset. Instead, it finds unique semantic regions of the considered shape. In \cref{fig:coseg_supp}, we compare our segmentation results to shape parts of the COSEG dataset. \ourmethod{} selects finer regions with high fidelity to the user's input. For example, a click on the horse's head segments only the head. The corresponding part in COSEG is coarser and includes both the head and neck. For the alien shape, all three antennas are regarded as one part. In contrast, \ourmethod{} marks only the antenna with the point click.
}

%\smallskip
%\noindent \textbf{View generalization.}
\paragraph{View generalization}
\cref{fig:view_supp} present additional examples of the view generalization of \ourmethod{}. Our method segments the whole body of the guitar, even though its back side is occluded from the click on the front. Moreover, the second click on the back side of the guitar's bridge cannot be seen together in any 2D view with the first click. Nevertheless, our method selects the complete 3D regions of the bridge and body.

\siga{
%\smallskip
%\noindent \textbf{3D consistency.}
\paragraph{3D consistency}
Different from SAM~\cite{kirillov2023segment}, our method operates natively in 3D. \ourmethod{} predicts the segmentation probability at the vertex level, or in other words, our result lives on the 3D surface of the mesh. Thus, it is 3D-consistent \textit{by construction}. In~\cref{fig:consistency}, we compare our segmentation projected to 2D with SAM's result for the projected shape views. Since SAM operates on each view independently, it yields inconsistent predictions. By contrast, in our case, any 2D projection originates from the same 3D segmentation and thus, our method is consistent across all views.
}

%\smallskip
%\noindent \textbf{Separability.}
\paragraph{Separability}
We analyze \ourmethod{}'s segmentation prediction by plotting the histogram of probability values for all the shape vertices for different clicks. Results were presented in \cref{fig:separability_supp} in the main paper. We observe that the histogram contains two populations: vertices with high segmentation probability, which correspond to the region for the clicked point, and the rest of the vertices with low probability. To further demonstrate the separability of the model's predictions, we use the Otsu Threshold \cite{ostu1979threshold} and binarize the probabilities. As the figure has shown, the region before and after thresholding remains almost the same, indicating that the segmented part can be easily separated from the shape.

\siga{
%\smallskip
%\noindent \textbf{Comparison.}
\paragraph{Comparison}
We present additional comparison results on the PartNet dataset in \cref{fig:partnet_supp}. While other methods struggle to segment the shape part corresponding to the click, our method succeeds. 
}

\begin{figure}
\centering
\newcommand{\pl}{-2}
% trim: left, bottom, right, top
\begin{overpic}[width=\linewidth, trim=400 -20 310 10, clip]{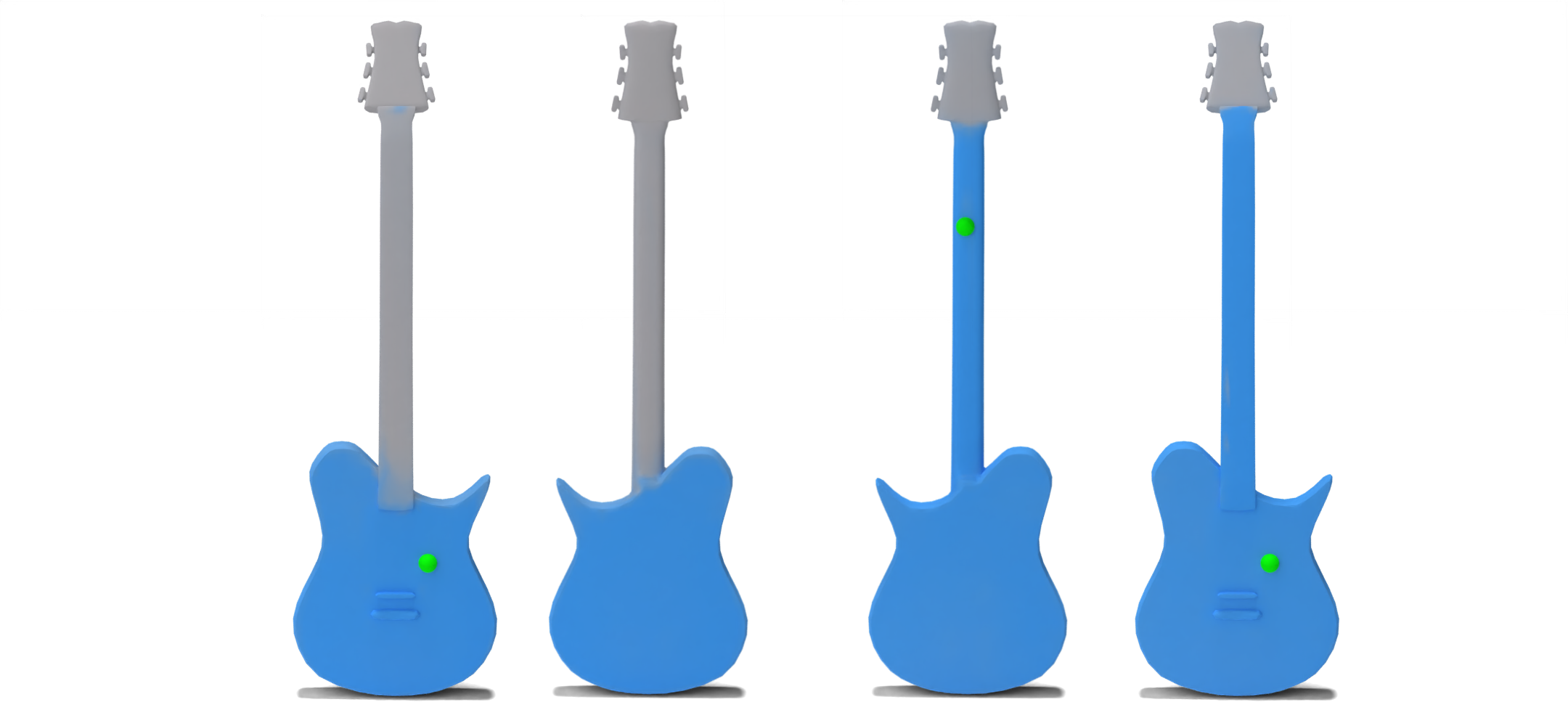}
\put(9.5,  \pl){\textcolor{black}{\small{First positive click}}}
\put(62.5,  \pl){\textcolor{black}{\small{Second positive click}}}
\end{overpic}
\vspace{-2mm}
\captionof{figure}{\textbf{View generalization.} Although our method was trained only with 2D supervision, it produces 3D segmentations for regions and clicks at opposite sides of the shape that cannot be seen together in 2D.}
\label{fig:view_supp}
\end{figure}

\begin{figure}[!t]
\centering
\newcommand{\pl}{-2}
% trim: left, bottom, right, top
\includegraphics[width=\linewidth, trim=100 0 0 0, clip]{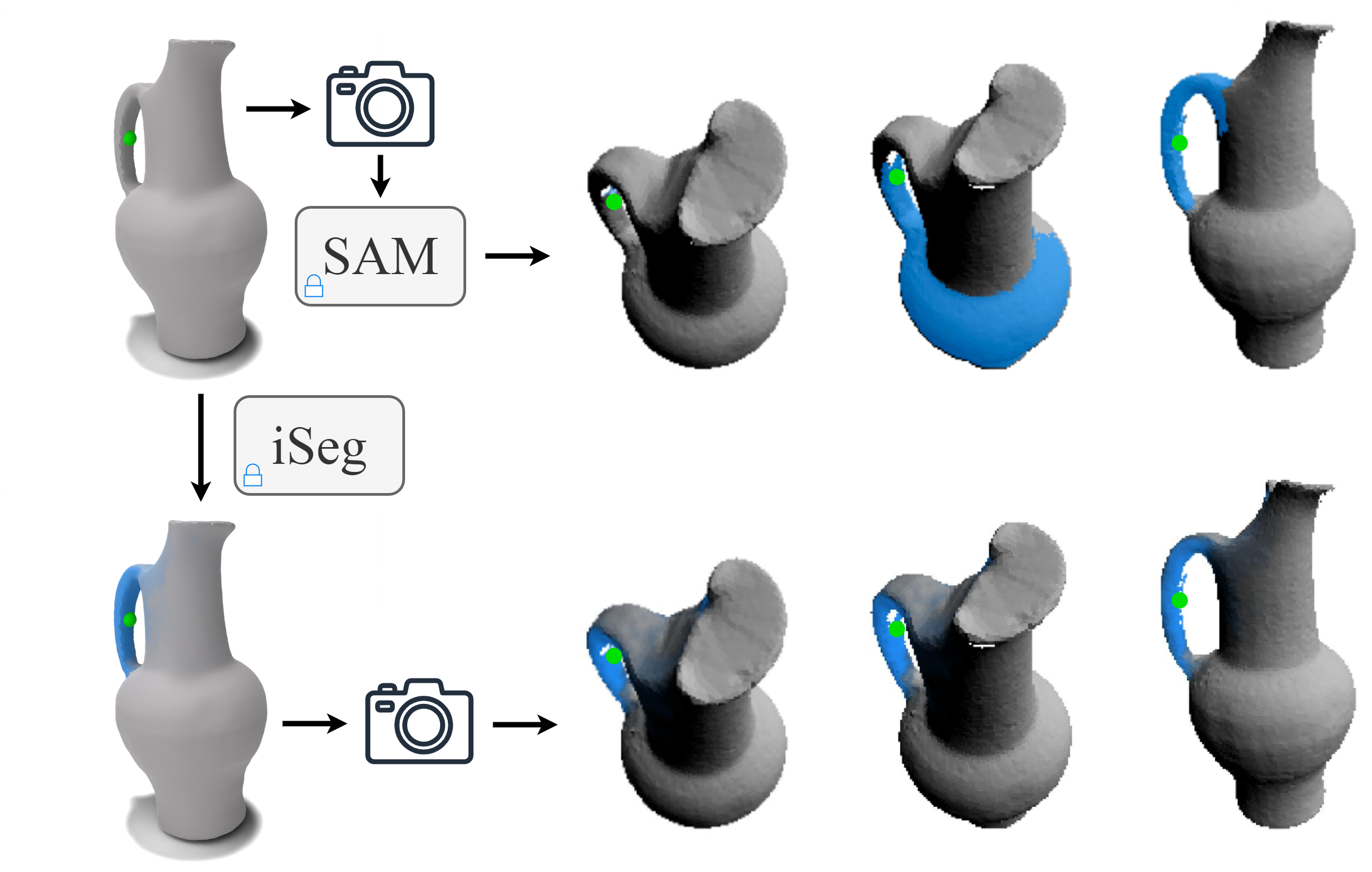}
\vspace{-5mm}
\caption{\siga{\textbf{Segmentation consistency.} SAM is highly sensitive to the viewing angle. It may generate substantially different masks for similar views, which are inconsistent in 3D. In contrast, \ourmethod{} is 3D consistent by construction.}}
\label{fig:consistency}
\end{figure}

\siga{
%\smallskip
%\noindent \textbf{Full-shape segmentation.}
\paragraph{Full-shape segmentation}
Our method's purpose is to select a customized shape region corresponding to user clicks. Nonetheless, we can easily employ \ourmethod{} for parsing the entire shape into multiple parts. In this setting, the user applies several interactive segmentation sessions. Assuming we have $K$ sessions, in each one of them we get the vertex segmentation probability $p_i^k$, where $i$ is the vertex index and $k = 1, ..., K$ is the session index. Then, the vertex class $c_i$ is determined according to the maximal confidence:

\begin{figure*}
\centering
\newcommand{\pl}{0}
% trim: left, bottom, right, top
\begin{overpic}[width=\linewidth, trim=165 0 255 120, clip]{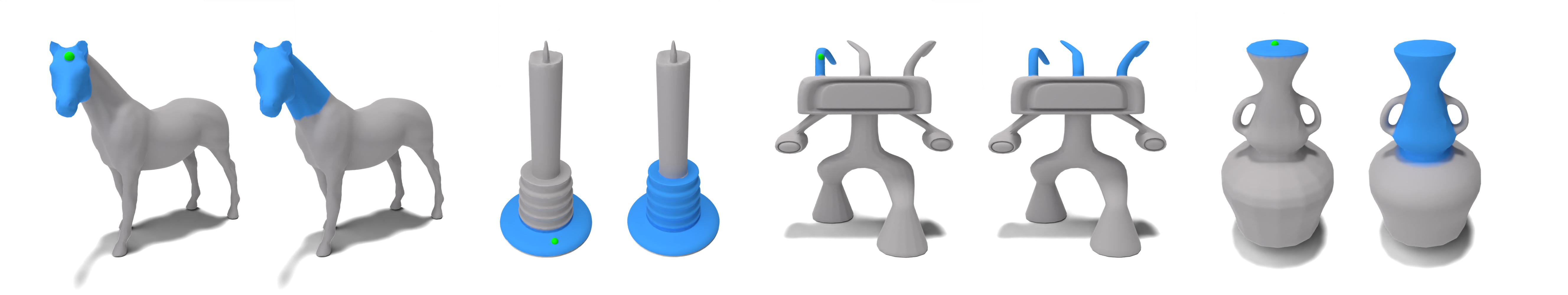}
\put(4.5,  \pl){\textcolor{black}{\small{\ourmethod{} result}}}
\put(18.5,  \pl){\textcolor{black}{\small{COSEG part}}}
\put(31,  \pl){\textcolor{black}{\small{\ourmethod{} result}}}
\put(39.5,  \pl){\textcolor{black}{\small{COSEG part}}}
\put(53.5,  \pl){\textcolor{black}{\small{\ourmethod{} result}}}
\put(67.5,  \pl){\textcolor{black}{\small{COSEG part}}}
\put(82,  \pl){\textcolor{black}{\small{\ourmethod{} result}}}
\put(91.5,  \pl){\textcolor{black}{\small{COSEG part}}}
\end{overpic}
\vspace{-5mm}
\captionof{figure}{\siga{\textbf{Segmentation fidelity.} Our method selects fine-grained regions with higher fidelity to the clicked point compared to the segment defined in the COSEG dataset \cite{coseg_2011} that contains the clicked point.}}
\label{fig:coseg_supp}
\end{figure*}

\begin{equation}
c_i =
\begin{cases}
\argmax\limits_{k} p_i^k, & \text{if $\max\limits_{k} p_i^k \geq t_{seg}$}.\\
0, & \text{otherwise}.
\end{cases}
\end{equation}

\noindent $t_{seg}$ is a threshold for assigning a vertex to one of the segmentation regions, and class ``0" is an additional background class. \ourmethod{} is typically confident in its predictions (with segmentation probabilities close to $0$ or $1$), thus, we set $t_{seg}$ to $0.5$.

\cref{fig:full_seg_supp} presents examples of complete shape segmentation with several single-click sessions. \ourmethod{}'s region prediction for clicks on different parts of the share are localized and mostly disjoint from each other. It enables a clear separation of the entire object into its unique parts without the need for any part-label supervision.
}

%\smallskip
%\noindent \textbf{Local geometric editing.}
\paragraph{Local geometric editing}
In \cref{fig:edits_supp} in the main body, we show how our method's results are used for local shape editing. \ourmethod{} selection corresponds to a contiguous region of an entire part of the shape. Thus, the segmented part can be easily manipulated. As the figure demonstrates, the entire bar of the chair is removed, only the head of the horse can be selected, the neck of the guitar is shortened, and a small local region of the candle's flame is enlarged.

\begin{figure}[!b]
\centering
% trim: left, bottom, right, top
\begin{overpic}[width=\columnwidth, trim=180 90 380 20, clip]{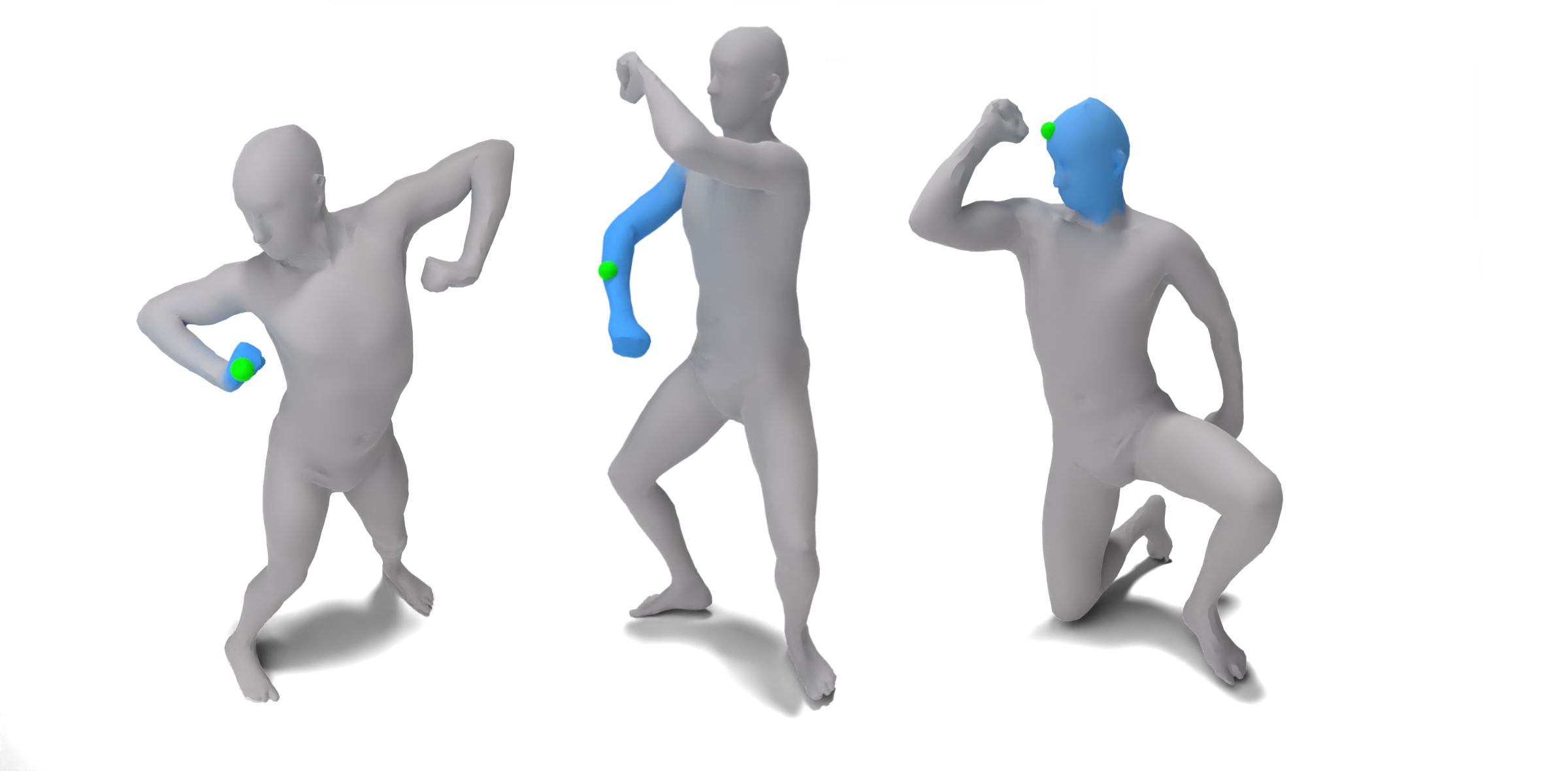}
\end{overpic}
\captionof{figure}{\camrdy{\textbf{Segmentaton of close body parts.} Although \ourmethod{} processes the mesh in the Euclidean space, it can separate body parts that are close in the Euclidean sense, but geodesic far.}}
\label{fig:humans_supp}
\end{figure}

\camrdy{
%\smallskip
%\noindent \textbf{Mesh processing in Euclidean space.}
\paragraph{Mesh processing in Euclidean space}
Our model operates on the vertex coordinates directly in Euclidean space without considering the shape topology explicitly. Such processing of the mesh, or surfaces in general, may potentially result in features that do not disambiguate surface points that are close in Euclidean space but are distant geodesically. Nonetheless, in practice, \ourmethod{} manages to disambiguate Euclidean nearby shape segments.

As demonstrated in \cref{fig:humans_supp}, when the human's hand or arm is close to the torso, or the head is close to the hand, only the part corresponding to the clicked point is selected. These results suggest that our method has learned different semantic features for different shape parts from the 2D foundation model \cite{kirillov2023segment}, enabling their separation. However, if the 2D supervision model does not disambiguate such regions successfully, \ourmethod{} may not distinguish them as well.
}

\siga{
%\smallskip
%\noindent \textbf{Stability.}
\paragraph{Stability}
A desired property of a click-based segmentation model is to be stable. That is, the results should be similar for nearby clicks that relate to the same shape part. This property implies that the model is robust to the exact click location and alleviates the need for precision from the user. \cref{fig:stability} demonstrates the stability of our method. As the figure shows, \ourmethod{} consistently selects the handle part of the hammer for different clicks along the handle.

We evaluate the stability property quantitatively by comparing the segmentation prediction for neighboring vertices. Given a clicked point, we compute the Intersection over Union (IoU) for its resulting mask and the mask for clicking each of its one-ring neighbors. We use 7 meshes with 100 random clicks from each mesh. For comparison, we utilize the SAM baseline, as described in \cref{sec:fidelity} in the main paper. The average IoU for \ourmethod{} is 0.8 compared to 0.7 for the SAM baseline. SAM's masks originate from 2D views of the shape. These predictions do not segment occluded regions of the shape and are less smooth in 3D. In contrast, our method operates directly in 3D and exhibits better stability than the SAM baseline.
}

In \cref{fig:stability_multi_supp}, we present the segmentation stability of \ourmethod{}. For a couple of clicks, the user can select the entire trailer box or only its side with various locations for the positive clicks or the positive and negative clicks, respectively.

%\smallskip
%\noindent \textbf{Granularity.}
\paragraph{Granularity}
\cref{fig:granularity_supp} exemplifies an additional property of our method: segmentation granularity. When segmenting the leg for an intricate shape like the human body, the user can control whether to exclude the foot, the lower shin part, or the upper part of the calf muscle by simply clicking the negative point at different locations along the leg.

\siga{
\section{Quantitative Evaluation}
\label{sec:quantitative_evaluation}

We provide additional details about our quantitative evaluation on the PartNet dataset \cite{mo2019partnet} discussed in \cref{sec:fidelity} in the main paper. The PartNet dataset includes meshes with ground-truth part labels. We measured the segmentation performance of \ourmethod{} and other methods by clicking a vertex and comparing the results to the ground-truth annotation of the part containing the click. \camrdy{We did not use any ground-truth information for training, but rather only for performance evaluation.}

For a robust evaluation, we randomly selected 5 vertices from the part \camrdy{that were not used during training}. Each vertex was regarded as a single-click interactive segmentation session, and we averaged the results over the clicked points and the evaluation shapes. \camrdy{We employed meshes from the dataset's test split for all the evaluated categories. According to our available resources, we used the 77 test shapes of the Knife category and sampled 93 additional meshes from the other categories.}

\begin{figure}[!tb]
\centering
\newcommand{\pl}{-2}
% trim: left, bottom, right, top
\includegraphics[width=\linewidth, trim=150 50 140 0, clip]{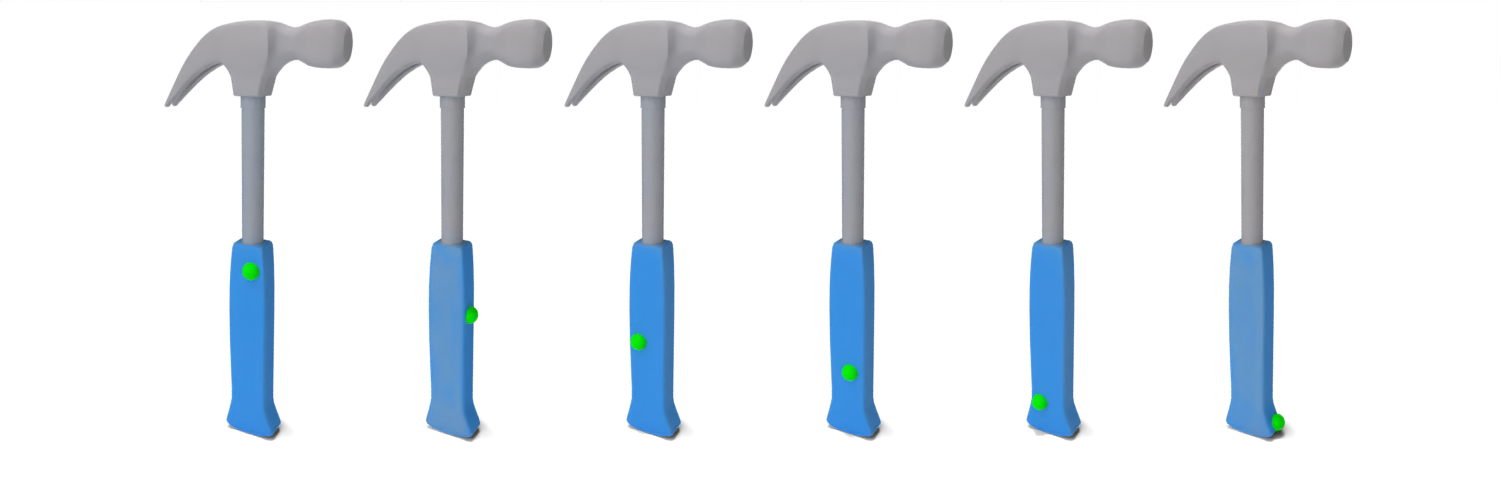}
\vspace{-5mm}
\caption{\siga{\textbf{Segmentaion stability.} The region selected by \ourmethod{} remains stable and consistent even when selecting different points within the same region.}}
\label{fig:stability}
\end{figure}

\subsection{Segmentaiton Binarization} \label{sec:seg_binarization}
\ourmethod{} predicts for each mesh vertex a soft segmentation value $p_i \in [0, 1]$. However, the part labels in the PartNet dataset are categorial. Thus, we binarized our prediction to obtain a per-vertex segmentation result $q_i \in \{0, 1\}$. The binarization was done with the Otsu Threshold \cite{ostu1979threshold}:

\begin{equation} \label{eq:binarization}
q_i = \mathbbm{1}(p_i \geq t_{otsu}),
\end{equation}

\noindent where $\mathbbm{1}{(\cdot)}$ is the indicator function and $t_{otsu}$ is the Otsu Threshold. We chose this simple binarization approach to minimize the post-processing effect and reflect our method's performance as faithfully as possible. For a fair comparison, we applied the same binarization approach to the other baselines in our evaluation.

\subsection{Evaluaiton Metrics} \label{sec:evaluaiton_metrics}
PartNet shapes are composed of mesh elements for each part with duplicate vertices at the elements' connection. To circumvent this duplication, we converted the vertex prediction to face prediction $s_i$, where the face prediction was considered as $``1"$ when all its vertices were segmented, and $``0"$ otherwise. To compute the Accuracy (ACC) and Intersection over Union (IoU) metrics, we labeled the part that included the clicked point as $``1"$, and the rest of the shape as label $``0"$. The ACC measured the rate of prediction $s_i$ that matches the face segmentation label $l_i$:

\begin{equation} \label{eq:seg_acc}
ACC = \frac{1}{n}\sum_{i = 1}^{n}{\mathbbm{1}(s_i = l_i)}.
\end{equation}

\noindent For computing the IoU, we divided the intersection of the segmentation prediction and the ground-truth annotated region by their union:

\begin{equation} \label{eq:seg_iou}
IoU = \sum_{i = 1}^{n}{(s_i \land l_i)} \; \Big/ \, \sum_{i = 1}^{n}{(s_i \lor l_i)}.
\end{equation}

\begin{figure}[!b]
\centering
\newcommand{\pl}{-2}
% trim: left, bottom, right, top
\begin{overpic}[width=\linewidth, trim=290 -50 190 0, clip]{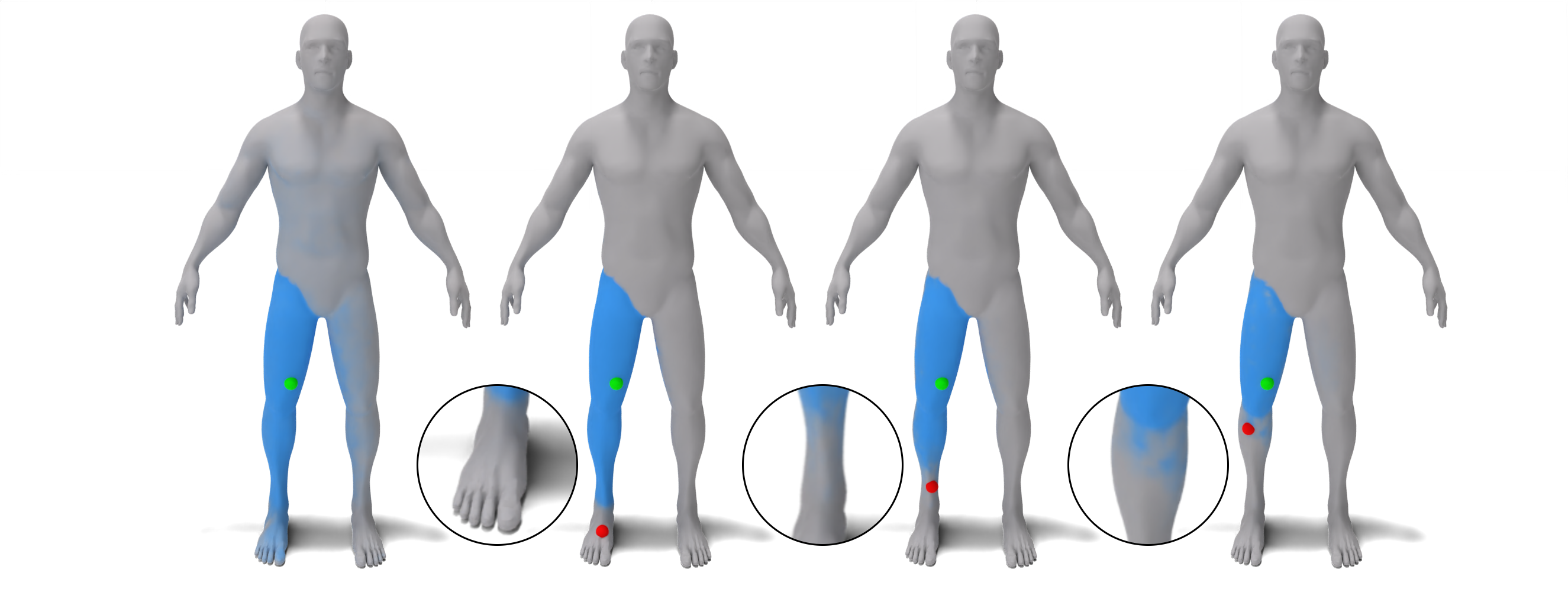}
\put(3,  \pl){\textcolor{black}{\small{1\textsuperscript{st} positive}}}
\put(26,  \pl){\textcolor{black}{\small{2\textsuperscript{nd} negative}}}
\put(52,  \pl){\textcolor{black}{\small{2\textsuperscript{nd} negative}}}
\put(77,  \pl){\textcolor{black}{\small{2\textsuperscript{nd} negative}}}
\end{overpic}
\captionof{figure}{\textbf{Segmentation granularity.} \ourmethod{} offers control over the segmented region according to the location of the clicked point.}
\label{fig:granularity_supp}
\end{figure}

\subsection{SAM Baselines} \label{sec:sam_baselines}

As explained in \cref{sec:fidelity} in the paper, we utilized the 2D segmentation model SAM \cite{kirillov2023segment} for segmenting shape parts. We evaluated several approaches to aggregate SAM's predictions into 3D. In the first method, we applied a union between SAM's masks (union aggregation). Meaning, a vertex belonged to the segmented region if it was segmented in one of the projected views of the shape. In the second method, we divided the number of times a vertex was segmented in the rendered views of the shape by the total number of rendered views (uniform average aggregation). In the third, instead of having the same division factor for all the vertices, we divided by the number of times a vertex was visible among the shape views (visibility average aggregation). \cref{tab:sam_baselines} presents the segmentation performance for our quantitative evaluation on PartNet, as described in \cref{sec:fidelity,sec:seg_binarization,sec:evaluaiton_metrics}. The visibility average aggregation had better results, thus, we used this baseline in the paper for comparison with our method.
}

\begin{table}[t!]
\caption{\siga{\textbf{Quantitative PartNet results for baselines using SAM.} We experimented with different methods of utilizing SAM for 3D part segmentation for shapes from the PartNet dataset. Considering SAM prediction averaged according to the vertex visibility (Visibility Average) yielded better results than the other options. Further details are given in \cref{sec:sam_baselines}.
}}
\vspace{-5pt}
\centering
\begin{tabular}{lccc}
\toprule
Aggregation  & Uniform Average & Union & Visibility Average \\
\midrule
Accuracy $\uparrow$ & 0.70 & 0.73  & 0.76 \\
IoU $\uparrow$ & 0.38 & 0.50 & 0.51 \\
\bottomrule
\end{tabular}
\vspace{-5pt}
\label{tab:sam_baselines}
\end{table}

\section{Perceptual User Study}
\label{sec:user_study}

In \cref{fig:effectiveness} in the main paper, we show segmentation results with different levels of effectiveness, as exemplified in our perceptual user study (discussed in \cref{sec:fidelity}). In the presented example, the point click is on the cubic region of an alien shape, and the segmentation result is visualized from different viewing angles, part of which the clicked point is occluded. 

The example in the top row is considered \textit{completely effective} with a score of 5 since the whole cubic region of the alien is segmented and no other region is selected. The result in the middle row is considered \textit{partially effective}, since not the whole cubic region is segmented (the back of the box is not selected), and a region other than the box is marked (part of the back leg). In the last row, the segmentation is considered \textit{completely ineffective} with a score of 1, since no region from the shape is selected (there is no blue color). The participants in our study were shown the segmentation results of each of the compared methods for each mesh and rated them according to the scoring explanation elaborated here.

\siga{\cref{fig:questions} presents several example questions from our perceptual user study, demonstrating the comparison between the alternative techniques.} The study's results, reported in \cref{tab:perceptual_study} in the paper, suggest that the segmented regions by \ourmethod{} are considered highly effective, compared to the partially effective segmentations by SAM baseline. This finding is consistent with the 2D nature of SAM. It is challenging to generate meaningful segmentation for occluded parts in 3D.

For example, in \cref{fig:sam_occlude} in the paper, the user's point click is on one side of the camel's hump. For a viewing angle where the click is not visible, the SAM baseline can only segment the hump partially. In contrast, \ourmethod{} fully segments the hump, even though the point click is concluded from the presented viewing angle. Due to \ourmethod{}'s 3D consistency, it can combine shape information from different viewing angles to produce an effective segmentation that is meaningful in 3D.

\section{Ablation Test}
\label{sec:ablation_tests}

We performed an ablation experiment of training the encoder and decoder together, rather than separately, as proposed in our method. We optimized the parameters of both the encoder and decoder (including the parameters of the interactive attention layer) together considering the loss of both the encoder and decoder. The loss function in the ablation test was: 
\begin{equation}
\Lcal = w \Lcal_{enc} + \Lcal_{dec},
\end{equation}
\noindent where $w = 5$ is the weight of the encoder loss. The definitions of $\Lcal_{enc}$ and $\Lcal_{dec}$ are the same as in the main paper: $\Lcal_{enc}$ is the $l_2$ loss between the projected mesh feature field and SAM's 2D embedding for the rendered shape image, and $\Lcal_{dec}$ is the binary cross-entropy loss between the resterized \ourmethod{} probability prediction and the reference 2D segmentation mask from SAM. We use a similar view generation scheme as in the main paper to train the ablation model.

We found that the ablation model and the original model proposed in the main paper perform equally well for the training vertices, as shown on the left side of \cref{fig:ablation} in the main paper. However, the ablation model suffers from over-fitting due to the high flexibility of the network when training the encoder and the decoder simultaneously, making it less powerful on the test vertices, as shown on the right side of \cref{fig:ablation}. \ourmethod{} generalizes better when trained for the same number of epochs as the ablation model.

\section{Implementation Details}
\label{sec:implementation_det}

%\noindent \textbf{Encoder.}
\paragraph{Encoder}
We use the vertex coordinates of the mesh and apply positional encoding of size 515, which is the input to our encoder. The encoder is implemented as a multi-layer perceptron (MLP) of 6 layers, with 256 neurons per layer. Each layer except the last includes ReLU activation and layer normalization. After the last layer, we apply hyperbolic tangent activation without normalization. The encoder's output is a per-vertex feature vector of size $256$. \siga{We trained the encoder with a wide range of random views from the full range $[-\frac{\pi}{2}, \frac{\pi}{2})$ for the elevation angle and $[0, 2\pi)$ for the azimuth angle. The resolution of the rendered image was $224 \times 224$ pixels to match the resolution of the SAM ViT-H model used in our experiments.}

%\smallskip
%\noindent \textbf{Decoder.}
\paragraph{Decoder}
The interactive attention module contains linear layers of size $256 \times 256$ for $W^Q$, $W_{pos}^K$, $W_{neg}^K$, $W_{pos}^V$, and $W_{neg}^V$. The output of this layer of size $256$ is concatenated to the per-vertex encoded feature vector of size $256$ to a total size of $512$. Then, a 16-layer MLP operates on the concatenated feature vector. The MLP has the first layer of size $512$, 14 layers of size $256$, and the last layer of size $2$. Each layer except the last includes ReLU activation and layer normalization. After the last layer, we apply a Softmax operation without normalization. \siga{As for the encoder, we used a wide range of random views for training the decoder (where the clicked points are visible), with the same $224 \times 224$ image resolution.}

\begin{figure*}[!t]
\centering
\newcommand{\pl}{-0.5}
% trim: left, bottom, right, top
\begin{overpic}[width=\linewidth, trim=50 200 50 100, clip]{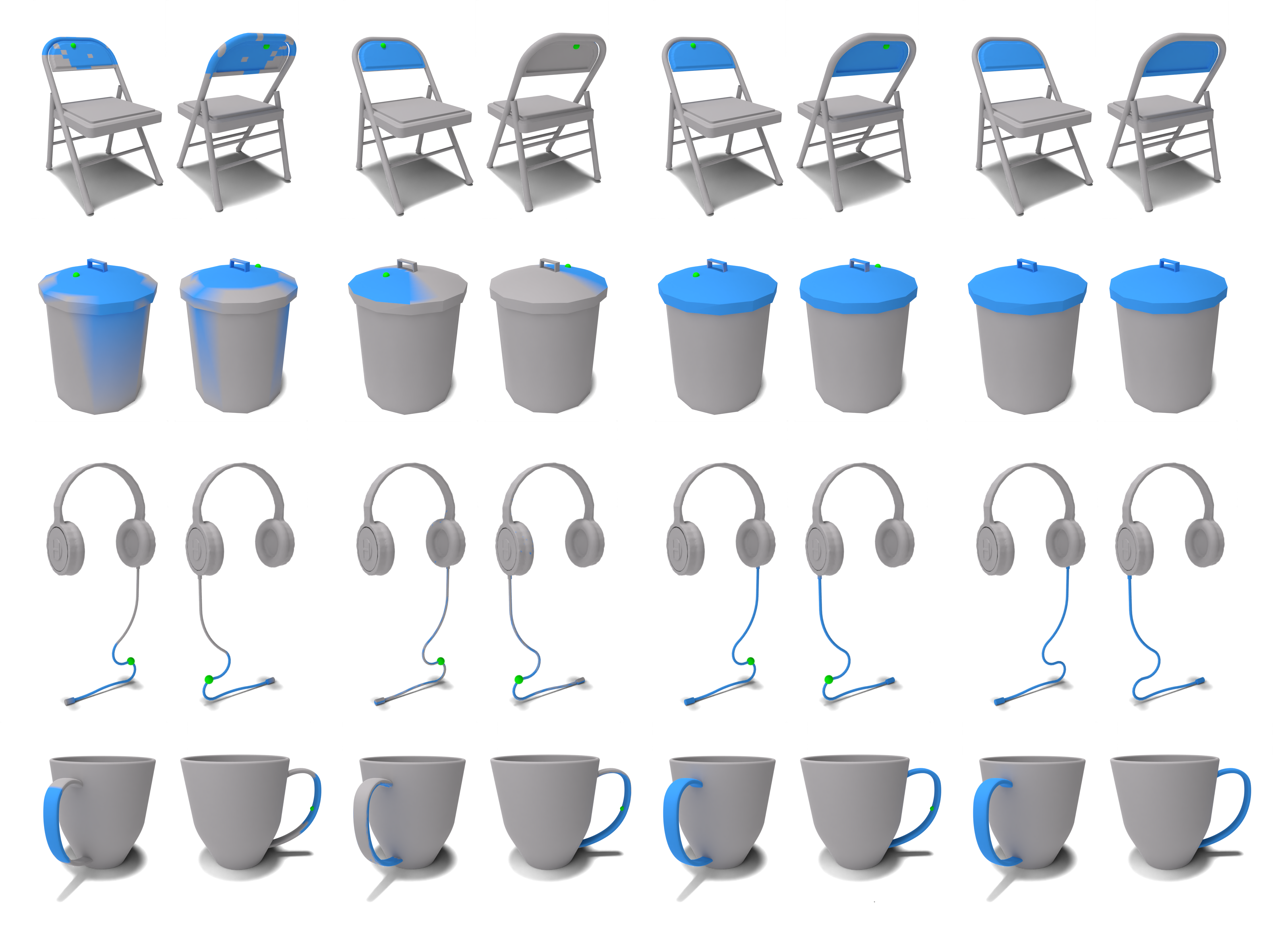}
\put(8,  \pl){\textcolor{black}{InterObject3D}}
\put(32.5, \pl){\textcolor{black}{SAM Baseline}}
\put(58, \pl){\textcolor{black}{iSeg (ours)}}
\put(82, \pl){\textcolor{black}{Ground-truth}}
\end{overpic}
\vspace{-5mm}
\caption{\siga{\textbf{Additional comparison results on the PartNet dataset.} Each pair shows different views of the segmentation result for different interactive techniques. \ourmethod{} produces better segmentation results, than the compacted methods, yielding segmentations that are close to the ground-truth part annotations.}}
\label{fig:partnet_supp}
\end{figure*}

%\smallskip
%\noindent \textbf{Compared method.}
\paragraph{Compared method}
As mentioned in the main body (\cref{sec:fidelity}), we compared the effectiveness of our method with the recent interactive segmentation work InterObject3D \cite{kontogianni2023interactive}. \camrdy{Since the training code was unavailable,} we adopted their publicly available pretrained model\footnote{\url{https://github.com/theodorakontogianni/InterObject3D/tree/main}} and used channels for point clicks, as instructed in their paper. For our experiments, we set the volume length for the point click to be 0.05.

\begin{figure*}[!t]
\centering
\newcommand{\plu}{51}
\newcommand{\pld}{-0.5}
% trim: left, bottom, right, top
\begin{overpic}[width=\linewidth, trim=20 -50 20 50, clip]{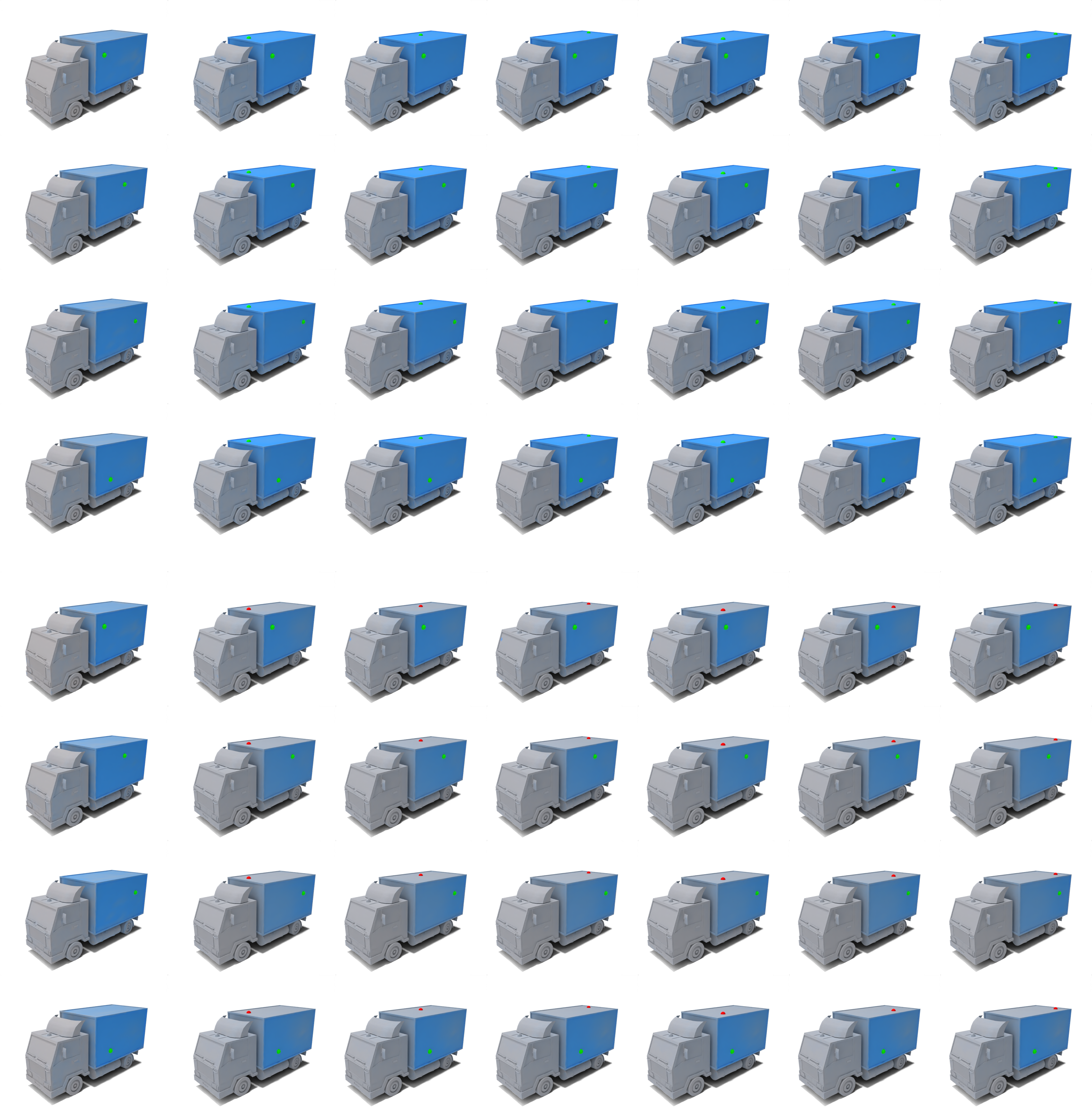}
\put(2.5,  \plu){\textcolor{black}{\small{First positive}}}
\put(17,  \plu){\textcolor{black}{\small{Second positive}}}
\put(30.5,  \plu){\textcolor{black}{\small{Second positive}}}
\put(44,  \plu){\textcolor{black}{\small{Second positive}}}
\put(57.5,  \plu){\textcolor{black}{\small{Second positive}}}
\put(71.5,  \plu){\textcolor{black}{\small{Second positive}}}
\put(85,  \plu){\textcolor{black}{\small{Second positive}}}
\put(2.5,  \pld){\textcolor{black}{\small{First positive}}}
\put(17,  \pld){\textcolor{black}{\small{Second negative}}}
\put(30.5,  \pld){\textcolor{black}{\small{Second negative}}}
\put(44,  \pld){\textcolor{black}{\small{Second negative}}}
\put(57.5,  \pld){\textcolor{black}{\small{Second negative}}}
\put(71.5,  \pld){\textcolor{black}{\small{Second negative}}}
\put(85,  \pld){\textcolor{black}{\small{Second negative}}}
\end{overpic}
\vspace{-3mm}
\captionof{figure}{\textbf{Segmentation stability for a couple of clicks.} The segmented region remains similar for positive and negative clicks with varying click locations over the semantic shape region.}
\label{fig:stability_multi_supp}
\end{figure*}

\begin{figure*}
\centering
\newcommand{\pl}{-2}
% trim: left, bottom, right, top
\includegraphics[width=0.955\linewidth, trim=0 0 0 0, clip]{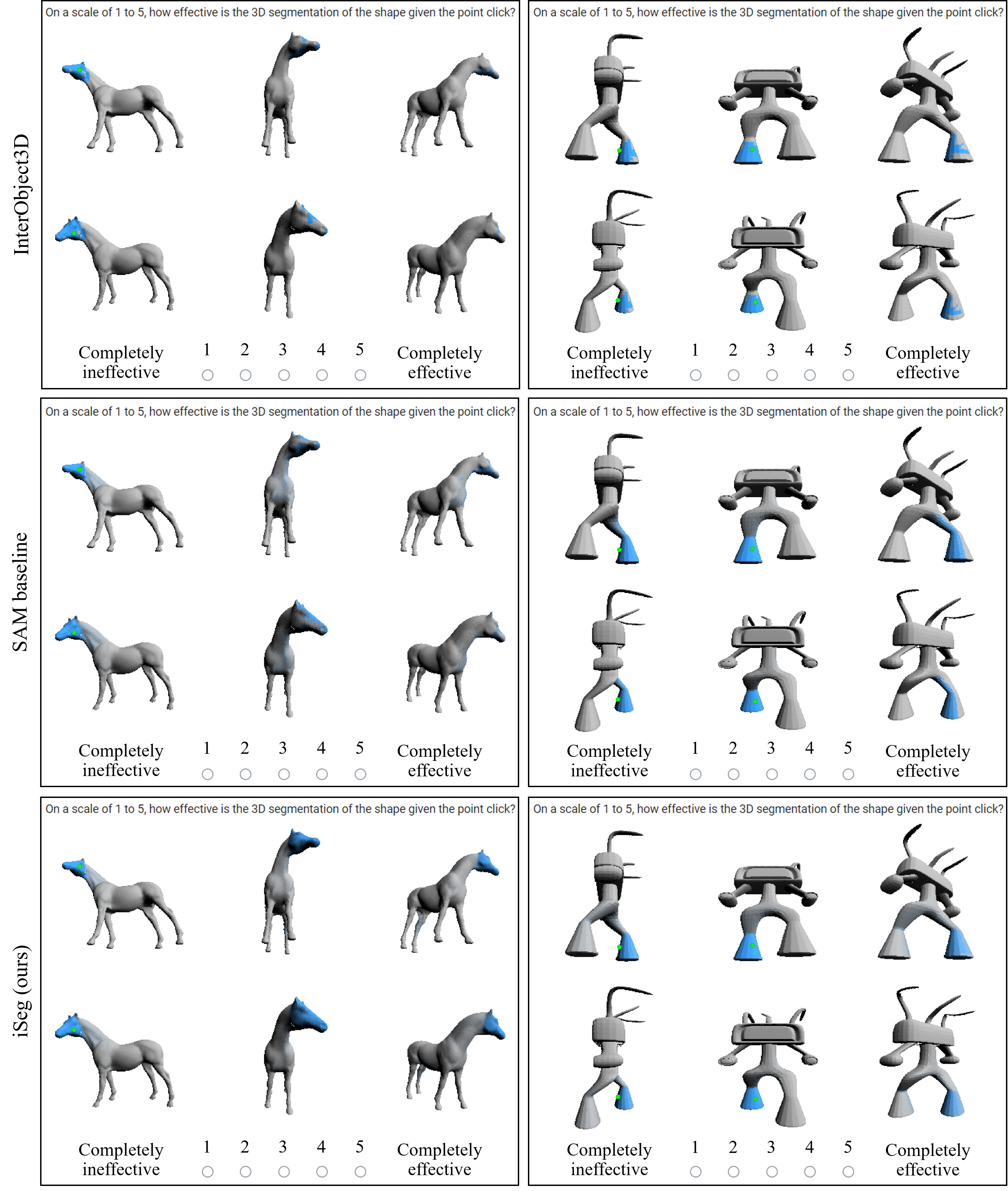}
\caption{\textbf{Perceptual study example questions.} For each mesh, we ask the participants to rate the effectiveness of the segmentation for the compared methods, where they are unaware of which method is used. The other techniques are perceived as partially effective, as they do not mark the entire 3D region corresponding to the clicked point. In contrast, \ourmethod{} selects complete regions in a 3D-consistent fashion and is considered highly effective.}
\label{fig:questions}
\end{figure*}

\fi

\end{document}